\documentclass{article} % For LaTeX2e
\usepackage{iclr2027_conference,times}
\iclrfinalcopy   % <-- de-anonymizes; comment out for the blind submission
\newcommand{\add}[1]{{\color{black}#1}}

\usepackage{hyperref}
\usepackage{url}

\title{Generation of High-Level Concepts in 3D Scene Graphs via Autoregressive Diffusion}

\author{
Jose Andres Millan-Romera\thanks{Equal contribution.} \\
Automation and Robotics Research Group, SnT\\
University of Luxembourg, Luxembourg\\
\texttt{jose.millan@uni.lu} \\
\And
Samuel Cognolato\footnotemark[1] \\
University of Padova, Padova, Italy\\
Fondazione Bruno Kessler, Trento, Italy $\quad$\\
\texttt{samuel.cognolato@unipd.it} \\
\AND
Holger Voos \\
Automation and Robotics Research Group, SnT,\\
and Faculty of Science, Technology and Medicine,\\
University of Luxembourg, Luxembourg\\
\texttt{holger.voos@uni.lu} \\
\And
Jose Luis Sanchez-Lopez \\
Automation and Robotics Research Group, SnT\\
University of Luxembourg, Luxembourg\\
\texttt{joseluis.sanchezlopez@uni.lu} \\
\AND
Luciano Serafini \\
Fondazione Bruno Kessler, Trento, Italy\\
\texttt{serafini@fbk.eu}
}

\usepackage{microtype}
\usepackage{graphicx}
\usepackage{subcaption}
\usepackage{booktabs} % for professional tables
\usepackage{algorithm}
\usepackage{algorithmic}
\usepackage{placeins}
\usepackage{hyperref}

\usepackage{amsmath}
\usepackage{amssymb}
\usepackage{mathtools}
\usepackage{amsthm}
\newtheorem{theorem}{Theorem}

\usepackage{acronym}
\usepackage{multirow}
\usepackage{adjustbox}
\usepackage{rotating}

\usepackage[table]{xcolor}

\usepackage{amssymb}

\usepackage[inline]{enumitem}
\usepackage{bm}

\usepackage{booktabs}
\usepackage{multirow, makecell}

\usepackage{todonotes}

\newlist{todolist}{itemize}{2}
\setlist[todolist]{label=$\square$}
\usepackage{pifont}
\newcommand{\cmark}{{\color{black}\checkmark}}
\newcommand{\xmark}{{\color{black}\ding{55}\ }}
\DeclareMathAlphabet{\mathsfit}{\encodingdefault}{\sfdefault}{m}{sl}
\SetMathAlphabet{\mathsfit}{bold}{\encodingdefault}{\sfdefault}{bx}{n}

\def\gG{{\mathcal{G}}}

\def\verts{{\mathcal{V}}}
\def\edges{{\mathcal{E}}}
\def\concepts{{\mathcal{C}}}
\def\planes{{\mathcal{P}}}

\def\fX{{\bm{X}}}       % X node features
\def\fE{{\bm{E}}}     % E edge features
\def\fR{{\bm{R}}}       % R real node features (e.g., coordinates)
\def\fr{{\bm{r}}}       % vector r
\def\fD{{\bm{D}}}     % D real edge features (e.g., distances)
\def\fZ{{\bm{Z}}}       % tensor graph
\def\fQ{{\bm{Q}}}       % matrix
\def\fpi{{\bm{\pi}}}
\def\fn{{\bm{n}}}

\def\R{{\mathbb{R}}}    % real numbers set
\def\N{{\mathcal{N}}}   % normal dist

\def\v0{{\mathbf{0}}}   % zero vector

\def\mI{{\bm{I}}}       % identity matrix

\newcommand{\CE}{\operatorname{CE}} % Cross Entropy
\def\topk{{\operatorname{topk}}}

\newcommand{\nlvalpm}[2]{\begin{tabular}[c]{@{}c@{}}#1 \\[-5pt] \scriptsize{$\pm$ #2}\end{tabular}}

\newcommand{\best}[1]{\cellcolor{gray!40}{\begingroup\bfseries\boldmath #1\endgroup}}
\newcommand{\second}[1]{\cellcolor{gray!18}{#1}}

\acrodef{3DSG}{3D Scene Graph}
\acrodef{IFH}{Insert-Fill-Halt}
\acrodef{FLAGG}{FLexible Autoregressive Graph Generation}
\acrodef{GNN}{Graph Neural Network}
\acrodef{D4}{Distance and Discrete Denoising Diffusion Model}
\acrodef{SLAM}{Simultaneous Localization And Mapping}
\acrodef{MiDi}{Mixed Graph and 3D Denoising Diffusion}

\acrodef{GAT}{Graph Attention Network}
\acrodef{GT}{Graph Transformer}
\acrodef{VAE}{Variational Autoencoder}
\acrodef{MDS}{Multidimensional Scaling}
\acrodef{FGW}{Fused Gromov-Wasserstein}
\acrodef{DGG}{Deep Graph Generation}

\begin{document}

\maketitle

\begin{abstract}
Indoor 3D Scene Graphs (3DSGs) enhance robot-observed geometric primitives, such as wall planes, with hierarchies of increasingly abstract metric-semantic concepts (walls, rooms, floors, buildings), each represented as a node with a semantic label, a 3D position, and links to the lower-level elements it groups. These hierarchies support robotic perception and Simultaneous Localization and Mapping (SLAM), yet inferring them automatically from the observed primitives remains an open problem: classical methods rely on rules hand-crafted for each concept, while existing learning-based methods use separate models for structure and for geometry, and do not generate beyond walls and rooms. We propose a unified graph generative model based on autoregressive diffusion that grows a complete 3DSG bottom-up from the observed planes: at each level, it jointly samples how many concept nodes to add, their labels and positions, and their connections to the level below, without any concept-specific module. To evaluate generated hierarchies, we adapt the Fused Gromov–Wasserstein distance to match generated and ground-truth nodes on both topology and features. Across synthetic scenes, real architectural floor plans, and robotic sensor data, up to city level, our model outperforms prior learning-based baselines and, on all the datasets but one, improves on a one-shot diffusion model even when the latter is given the ground-truth graph size, with the largest gains on the deepest hierarchies and on robot-recorded environments.
\end{abstract}
\section{Introduction}

\begin{figure}[t]
  \begin{center}
    \centerline{\includegraphics[width=\columnwidth]{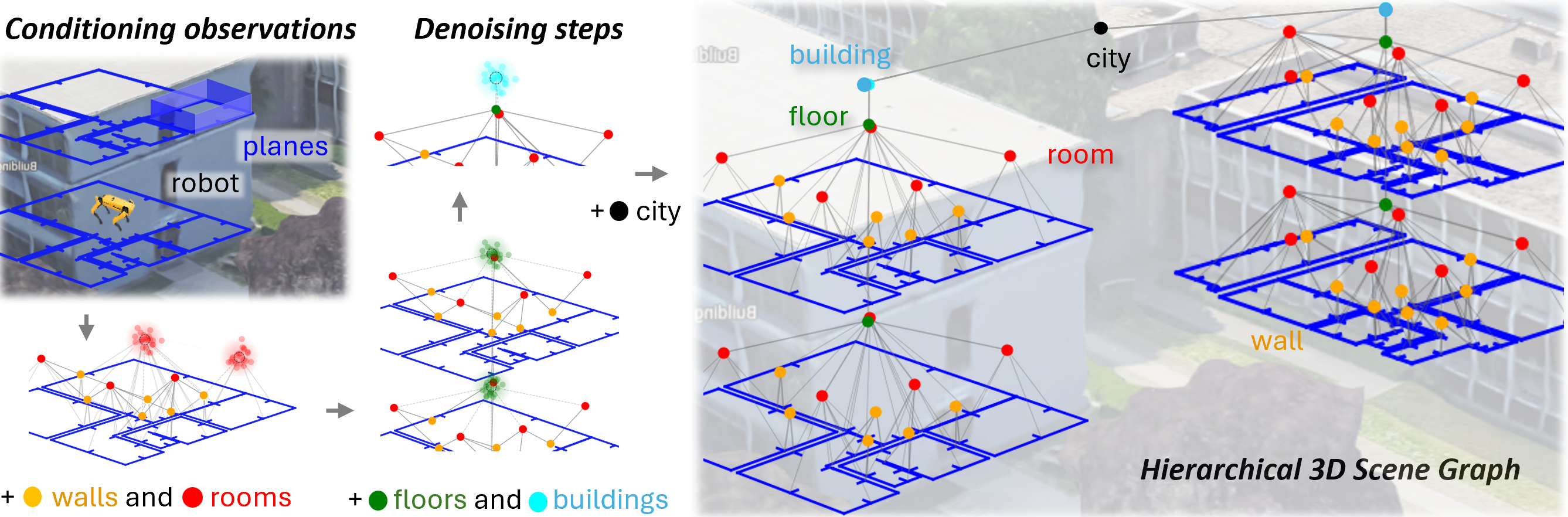}}
    \caption{
      Hierarchical 3D Scene Graph Generation. Starting from wall planes observed by the robot (blue lines), levels are generated bottom-up, with nodes colored by class and placed at their centroids. Grey edges denote part-of relations.
      }
    \label{fig:front}
  \end{center}
\end{figure}

Robots operating in complex indoor environments must reason about spatial structure at multiple scales, from individual sensor measurements to building-wide topology. \acp{3DSG} in robotics~\citep{3d_scene_graph, hydra} have emerged as a compact representation of this structure, combining semantic, geometric, and relational information. As illustrated in Fig.~\ref{fig:front}, the hierarchy comprises (i) an observed, low-level layer of planes measured directly from sensor data, and (ii) higher-level layers that aggregate lower-level elements into increasingly abstract topological concepts (e.g., rooms). Formally, a \ac{3DSG} is a graph: every plane and every concept is a node carrying a semantic label and a 3D centroid, and edges encode part-of relations that link each concept to the lower-level nodes it aggregates. These concepts, together with their estimated centroids, enhance downstream robotic applications such as \ac{SLAM}~\citep{bavle2022s}, global localization~\citep{shaheer_deviations}, path planning~\citep{ejaz2025situationally}, and multi-robot communication~\citep{fernandez2024multi}.

% These concepts, together with their estimated 3D poses or centroids, support downstream tasks such as \ac{SLAM}~\cite{bavle2022s}, global localization~\cite{shaheer_deviations}, path planning~\cite{ejaz2025situationally}, and multi-robot communication~\cite{fernandez2024multi}.

\add{Despite extensive work on predicting relations between observed nodes~\citep{wu2021scenegraphfusion}, inferring complete multi-level hierarchies from low-level observations remains largely open. Classical pipelines hand-craft one algorithm per concept, mostly rooms~\citep{luperto2022robust,hydra}, while existing learning-based methods stop generation at walls and rooms, split structure and geometry across separate models, or are restricted in layout complexity~\citep{millan2025metric}.}

% To overcome these limitations, we propose a unified approach for generating complete, multi-level \acp{3DSG} directly from an observed plane graph, removing the need for ad-hoc concept definitions, building on \ac{DGG}~\citep{cognolato2026flagg,vignac2023midi}. Our model learns a distribution over hierarchical scene graphs and generates new structures by progressively extending the observed graph bottom-up. We consider a static environment and an offline setting: the model takes the final plane map built by the robot and generates the full hierarchy from it in a single pass. Our approach jointly generates graph structure (nodes and edges) alongside semantic labels and continuous geometric attributes (centroids), conditioned on lower hierarchy levels. This yields consistent high-level structures, including walls, rooms, floors, buildings, and city-level entities.
% To evaluate the resulting graphs holistically, we introduce an adaptation of the \ac{FGW} metric~\cite{vayer2020fgw} that jointly accounts for semantic, metric, and relational fidelity.

To overcome these limitations we propose a unified model based on \ac{DGG}~\citep{cognolato2026flagg,vignac2023midi} that generates complete, multi-level \acp{3DSG} directly from the graph of observed planes, without ad-hoc concept definitions. The model learns a distribution over hierarchical scene graphs and, at inference, extends the observed graph bottom-up: at each level it adds the new concept nodes together with their labels, centroids, and edges to the level below, conditioned on the levels already generated. This yields consistent hierarchies spanning walls, rooms, floors, buildings, and cities. We consider a static environment and an offline setting: the input is the final set of planes mapped by the robot, and the full hierarchy is generated from it once. To evaluate the resulting graphs holistically, we adapt the \ac{FGW} metric~\citep{vayer2020fgw} to jointly account for semantic, metric, and relational fidelity.

Our key contributions are as follows:
\begin{enumerate*}[label=(\arabic*)]
    \item To the best of our knowledge, we present the first autoregressive, diffusion-based \ac{DGG} model for \ac{3DSG} completion, generating multi-level hierarchies conditioned on graphs of robot-observed planes.
    \item We introduce an adaptation of \ac{FGW} to evaluate hierarchical \ac{3DSG} generation via structure--attribute graph matching, jointly capturing relational, semantic, and metric consistency.
    \item We evaluate on diverse synthetic and real-world scenarios, showing higher accuracy and broader hierarchical coverage than prior methods, and we publicly release the datasets and code to facilitate benchmarking and future research.
\end{enumerate*}
\section{Related Work}
\label{sec:related_works}

\textbf{Graph generation.}
The generation of structured data in the form of graphs has been applied to drug design~\citep{zang2020moflow},  material design \citep{maziarka2020mol}, and social network analysis \citep{grover2019graphite}. The goal of graph generation is to approximate an unknown data graph distribution $q(G)$, of which a dataset is available, with a learned model $p_\theta(G)$, and sample new graphs from it with the same characteristics.
%Methods tackling the task comprise Variational Autoencoders~\cite{simonovsky2018graphvae}, Normalizing Flows~\cite{zang2020moflow},
State of the art methods include Diffusion Models~\citep{vignac2023digress} and Autoregressive Models~\citep{kong2023autoregressive}. There have been efforts to extend graph generation to multimodal graphs, containing both discrete labels and 3D positions. For 3D attributed graphs, methods must ensure SE(3) invariance, i.e., the probability distribution should be invariant to rotations and translations. This has been addressed through invariant features~\citep{vignac2023midi} or distance-based generation~\citep{cognolato2026d4}. The same invariance must be carried over for 3DSG generation, but these methods are not designed for extending existing graphs. The solution adopted in this paper is to apply graph autoregressive models, generating graphs as a sequence of node and edge insertions, incorporating the generation of 3D positions, labels and edges in each step.

\textbf{Generation of high-level concepts in 3DSGs.}
%The problem of capturing \emph{higher-level} concepts as properties that arise only when multiple observed concepts are considered jointly was first tackled with ad hoc robotic pipelines tailored to the target concept and the available sensor modality.
%A prominent example is the notion of \emph{rooms}: early works inferred room layouts from 2D occupancy maps~\cite{luperto2022robust}, and later incorporated them into \acp{3DSG} via 3D free space clustering~\cite{hydra}. Similarly, \cite{Bavle2025SGraphs2} introduces floor nodes in multi-story buildings through staircase detection.
Classical 3DSG pipelines rely on concept-specific heuristics tied to a
sensor modality: room segmentation from occupancy maps~\citep{luperto2022robust} or free space clustering~\citep{hydra}; and floor detection via geometric cues~\citep{Bavle2025SGraphs2}.
While effective, these pipelines typically encode concept-specific heuristics that are difficult to reuse across different high-level concepts.
The first learning-based approach considering these properties at the graph level was Neural Trees~\citep{talak2021neural}, which performs node classification via message passing on an auxiliary tree whose nodes correspond to subgraphs of the original graph, enabling hierarchical aggregation of set-level structure.
Other works focused only on the definition of relations between observed objects~\citep{wu2021scenegraphfusion, yang2017support}.
%\cite{millan2024learning} generates wall and room nodes by classifying semantic relations among plane primitives, but relies on a separate \ac{GAT} per concept type, assumes rectangular layouts, and does not predict centroids for newly generated nodes.
%\cite{millan2025metric} unifies the edge classification models and improves the grouping of planes into sets, additionally estimating centroids with a separate \ac{GAT}. Nevertheless, these pipelines still depend on intermediate edge classification and classical community detection, use distinct architectures for relation and centroids inference, and remain limited to a narrow set of generated node classes.
For node generative approaches, \citet{millan2024learning} introduce learned high-level node generation by classifying semantic relations between plane primitives to infer walls and rooms, but require separate \ac{GAT} models per concept type, assume rectangular layouts, and do not predict geometric attributes (centroids) for generated nodes, which are instead computed by a hand-crafted function from the grouped planes.
\citet{millan2025metric} unify edge classification into a single model and add centroid estimation via an additional \ac{GNN}\add{, handling non-rectangular rooms but requiring every wall to consist of exactly two planes.}
However, both methods share fundamental limitations: they depend on intermediate edge classification followed by classical community detection algorithms, employ separate architectures for structural and geometric inference, and generate only two node types (walls and rooms), preventing extension to multi-level hierarchies.
In contrast, our unified model jointly generates structure and centroids across arbitrary depths without concept-specific modules. 
%; thus, the literature lacks a unified framework that can \emph{jointly} generate a broad range of high-level node types together with their graph structure and continuous attributes, such as centroids, without relying on ad hoc heuristics or intermediate representations.

\section{Problem definition}
\label{sec:back/problem}

% \subsection{Notation}\label{sec:back/notation}
We represent sets with calligraphic letters ($\mathcal{S}$); scalars with lower case letters ($s$); vectors with lower case bold letters ($\bm{v}$); matrices and tensors with capital bold letters ($\bm{M}$). We define graphs as tuples $\gG=(\verts,\edges)$, where $\verts=\{v_1,\dots,v_n\}$ is the set of nodes and $\edges\subseteq\verts\times\verts$ is the set of edges. We denote functions on nodes as $f:\verts\to\mathcal{D}$, where $\mathcal{D}$ is the codomain, e.g., the set of labels in case $f$ is a node labeling function. Given a 3D node centroid function $\rho:\verts\to\R^3$, we define the node 3D Euclidean distance $d:\verts\times\verts\to\R_+$ as $
    d(v,v')=\|\rho(v)-\rho(v')\|
$
where $\|\cdot\|$ is the Euclidean norm.
%

% \subsection{Problem definition}\label{sec:back/problem}
We represent the 3D structure of a man-made indoor environment, e.g., a building, as a set $\planes$ of 3D vertical planes, each of the form $\fpi=(\fn,l,\fr)$, where $\fn\in\R^3$ is the 3D normal orientation, $l\in\R_+$ is its non-negative width and $\fr\in\R^3$ is the 3D centroid of the plane~\citep{millan2024learning}. These planes are typically detected by the robot using a LiDAR sensor or RGB-D camera, employing \ac{SLAM} (S-Graphs method~\citep{bavle2022s}) to filter outliers and refine plane estimates.
%to make more informed decisions in downstream tasks like path planning~\cite{ejaz2025situationally} or global localization~\cite{shaheer_deviations}.
%To enable... {\color{red}(For JA: please add details, what I mean here are: why are we interested in this in the first place)}, it is interesting to find relations between observed objects~\citneeded.
%The \ac{3DSG} is composed of low-level (observed) nodes and high-level nodes connected by edges that typically represent belonging.
% Topological concepts can be synthesized into higher-level spatial concepts whose semantic, metric, and relational properties depend on lower-level representations, forming a hierarchical \ac{3DSG}.
The  hierarchical \ac{3DSG} is defined bottom-up in the following order: \textit{plane}, \textit{wall}, \textit{room}, \textit{floor}, \textit{building}, \textit{city}, where most of the layers present a tree-like structure. Each high-level concept node directly depends on its children in the layer below. The exception is the wall layer, which connects to planes but has no parent node, breaking the strict hierarchy, a design choice consistent with other works in the field~\citep{millan2025metric}.
%These relations are expressed by connecting planes to higher-level concepts, representing for example rooms, walls, buildings, etc... . We include all concepts, both high-level and planes, in the set of concepts $\concepts$. The ones considered in this work are \textit{plane}, \textit{wall}, \textit{room}, \textit{floor}, \textit{building}, \textit{city}.
% Thus, the objective is to create a hierarchical representation of the 3D environment, connecting planes with higher-level concepts. % in a tree-like fashion.
%This structure, which we identify as 3D scene graph, is a 
\acp{3DSG} are undirected labeled graphs $G=(\gG,\lambda,\rho,\pi)$, where $\gG$ is the undirected graph topology; $\lambda: \verts \to \concepts$ is the 
function labeling nodes with their respective concepts $c\in\concepts$; $\rho: \verts \to \R^{3}$ is the 3D centroid function, returning coordinates $\fr$ for each node; $\pi: \verts \to \R^{3}\times\R_+$ is the plane features function, returning the plane normal vector and width $(\fn, l)$ for plane nodes, and a vector of zeroes otherwise. We define \ac{3DSG} generation as the task of inferring the correct \ac{3DSG} $G$ with high-level concepts given the set of all planes $\planes$ observed by the robot as evidence. Formally, we seek to train a conditional generative model $p_\theta(G|\planes)$ that approximates the dataset distribution $q(G|\planes)$. We assume a static environment and an offline setting: $\planes$ is the final, converged set of planes produced by the SLAM back-end after mapping is complete, and the 3DSG is generated once from it. Neither the scene nor the observations change during generation.

\section{Generating 3DSGs}
\label{sec:generating_3dsg}

\begin{figure*}[t!]
  \begin{center}
    \centerline{\includegraphics[width=\textwidth]{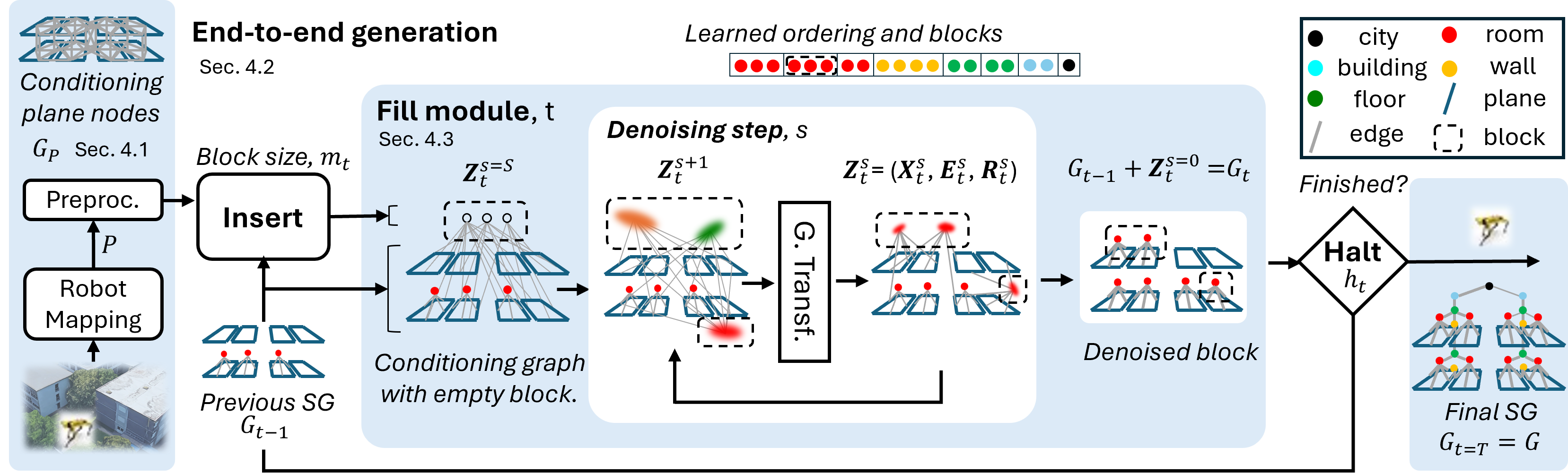}}
    % \caption{
    %   Conditioning plane nodes from robot mapping are first preprocessed (left), after which the model performs an iterative end-to-end generation of the complete 3DSG (right). At generation step $t$, the \textit{Insert} module selects the size of the next node block to be added to the previous graph $G_{t-1}$, while a GNN initializes node embeddings. The \textit{Fill} module denoises node classes, the adjacency matrix, and pairwise edge distances over $s$ and $r$ diffusion steps; the distances are converted to node positions via MDS, yielding the denoised block $G^{s=S}{t}$. Finally, the \textit{Halt} module predicts when the 3DSG is complete, producing $G_{t=T}$. The top row illustrates an example of the block ordering jointly learned by the modules for the depicted 3DSG. \add{The Fill module is depicted for the D4-based variant (\textsc{Ours-FD4}); the \ac{MiDi}-based variant replaces the two denoising stages and MDS with joint denoising of labels, edges, and coordinates (Sec.~\ref{sec:method/midi_adapt}).}
    %   \vspace{-0.6cm}
    % }
    \caption{Plane nodes from robot mapping are preprocessed into a proximity graph $G_\mathcal{P}$ (left), from which the 3DSG is grown iteratively (right). At generation step $t$, \textit{Insert} selects the size $m_t$ of the next block added to $G_{t-1}$; \textit{Fill} jointly denoises its classes, edges, and 3D positions from $Z^{s=S}_t$ to $Z^{s=0}_t$; and \textit{Halt} predicts $H_t$, ending generation at $G_{t=T}$.}
\label{fig:architecture}
  \end{center}
\end{figure*}

We model the hierarchy generation as the process of growing a graph, repeatedly adding new nodes and edges to the starting planes. The overall architecture is shown in Fig.~\ref{fig:architecture}. First, we transform the input set of 3D planes $\mathcal{P}$ into nodes, adding auxiliary edges based on their proximity (see Section~\ref{sec:method/init}). Then, the learned generative process proceeds iteratively: at each iteration, 
new nodes are introduced, after which their labels $\lambda$ in $\concepts$ (e.g., room, floor), 
connectivity $\edges$, and positions $\rho$ are sampled. The model then decides whether to halt 
or continue with a new iteration. The growing process is implemented by the \ac{FLAGG} framework~\citep{cognolato2026flagg}, while the connectivity and attributes are generated by a one-shot model inside \ac{FLAGG}; Fig.~\ref{fig:architecture} depicts the pipeline with the MiDi-based Fill module (\textsc{Ours}). We refer to Sec.~\ref{sec:method/flagg} and Sec.~\ref{sec:method/filler}, respectively, for the details. Our choice of an autoregressive model stems from two main reasons: (1) they are capable of efficiently adding new nodes and edges starting from a pre-existing graph; (2) they have been shown to require less memory than one-shot models~\citep{liao2019gran,cognolato2026flagg}, making 3DSG generation feasible. After the halting signal, the graph is returned, and the initial proximity edges are removed for the final result. 

\subsection{Initializing the graph of planes}\label{sec:method/init}
As the first step of our algorithm, we augment the input set of planes with 
edges based on their positions, building a graph. The resulting structure 
provides the generative model with relational context, which is then exploited 
by Graph Neural Network layers~\citep{scarselli2008graph,corso2024graph}. Formally, for each set $\planes$, we construct a graph $G_\planes$ with 
$\verts=\{v_1,\dots,v_{|\planes|}\}$, $\lambda$ always returning the 
\textit{plane} label, $\rho$ returning the 3D centroids, $\pi_\planes$ returning the correct plane features, and edges $\edges_\planes$ included through the following criterion:
\begin{equation}
    \edges_\planes = \left.\Big\{(v,v')\in\verts\times\verts \middle\vert v'\in\topk_\verts(-d(v,\cdot)),v\neq v'\Big\}\right.,
\end{equation}
where $\topk_\verts(-d(v,\cdot))$ returns the nearest $k$ nodes to $v$ based on the Euclidean centroid distances.

\subsection{FLAGG for graph extension}\label{sec:method/flagg}
Following the preprocessing step of the previous section, we formulate the sampling of the hierarchy $G$ from the planes as $p_\theta(G|G_\planes)$, i.e., the model extends the graph of planes $G_\planes$, adding edges, new high-level concept nodes and their centroids. To do so, we propose the use of \ac{FLAGG}, which enables the design of autoregressive models using modular components. At each step $t$, \ac{FLAGG} adds a block of new nodes and edges to the current graph $G_{t-1}$, producing a larger graph $G_t$. This yields a sequence $G_{0:T}=(G_0,\ldots,G_t,\ldots,G_T)$ of increasingly large graphs, with $G_0=G_\planes$ and $G_T=G_\text{final}$. The total number of steps $T$ is not fixed but determined at generation time by a binary halting signal $H_t$: generation continues until $H_t=1$. Each new graph $G_t$ at generation step $t$ is sampled from the transition probability $p(G_t,H_t|G_{t-1})$, defined as follows:
%\ac{FLAGG} samples across a variable amount of steps $T$, forming a sequence $G_{0:T}=(G_0,\ldots,G_t,\ldots,G_T)$ of increasingly larger graphs, with $G_0=G_\planes$, and $G_T=G_\text{final}$. The halting time $T$ is chosen when the binary halting signal $H_t$, returned at each step, is $H_t=1$. Each new graph $G_t$ at time $t$ is sampled from the transition probability $p(G_t,H_t|G_{t-1})$, defined as follows:
\begin{enumerate*}[label=(\it\roman*)]
    \item first, the \textit{Insertion model} $p^I_\phi(m_t|G_{t-1})$ samples the size $m_t$ of the next \textit{block}, i.e., the number of new nodes to be jointly added to the graph;
    \item the connectivity and any relevant feature, e.g., labels and centroids, of the new nodes are sampled by the \textit{Fill model} $p^F_\theta(G_t|m_t,G_{t-1})$, producing the next graph $G_t$;
    \item the end of generation is signaled by a positive sample from the \textit{Halting model} $p^H_\nu(H_t=1|G_t)$.
\end{enumerate*}
Formally, a single transition and the full distribution of \ac{FLAGG} are defined as:
\begin{gather}\label{eq:flagg_transition}
        p_{\phi,\theta,\nu}(G_t,H_t=h|G_{t-1}) = p^H_\nu(H_t=h|G_t) p^F_\theta(G_t|m_t,G_{t-1})p^I_\phi(m_t|G_{t-1}), \\
        p_{\phi,\theta,\nu}(G|G_\planes) = \sum_{T=1}^{\infty } \sum_{G_{0:T} \in \mathcal{I}(G,G_\planes,T)} \prod_{t=1}^{T} p_{\phi,\theta,\nu}(G_t,H_t=h_t|G_{t-1}),\label{eq:flagg_full}
\end{gather}
where $\mathcal{I}(G,G_\planes,T)$ is the set of all insertion sequences of graphs $G_{0:T}$, and the halting signal $h_t$ is $1$ for $t=T$, and $0$ otherwise.
We implement the Insertion and Halting models as Graph Isomorphism Networks with Edges (GINE)~\citep{Hu*2020Strategies}, predicting the block size $m_t$ and the halting signal $h_t$, respectively. The Fill model can be any one-shot graph generative model, adapted as in~\citet{cognolato2026flagg}. In the following section, we detail our choice of the Fill model. The \ac{FLAGG} modules are trained to recover nodes gradually removed from a data graph $G$ through a node removal process, stopping at $G_\planes$. This process characterization, comprising how many nodes are removed at each step and their order, directly influences the learned insertion policy, as it tries to reverse it. Thus, the node schedule $m_t$ and the node ordering $\sigma:\verts\to\{1,\ldots,|\verts|\}$ are two fundamental hyperparameters of \ac{FLAGG}, which can be designed taking into consideration the prior knowledge on the problem. For the node schedule, we use the efficient categorical removals from FLAGG which, given a fixed set $\mathcal{D}$ of possible block sizes, computes the smallest multiset of block sizes that sum to the number of nodes of $G$ minus those in $G_\planes$, and samples its elements in random order. We also devise a node order that considers the hierarchy of the graph, and refer the reader to Appendix~\ref{sec:method/order}.

\subsection{Generating edges, labels and centroids}
\label{sec:method/filler}
At each step of \ac{FLAGG}, the Fill module receives as input the current partial graph $G_{t-1}$ and the number of nodes to add $m_t$, and samples the edges $\edges_t$ across new nodes and old nodes, node labels $\lambda_t$, and centroids $\rho_t$ of the new nodes. At the end of sampling, the new features and edges are added to $G_{t-1}$ to compose the next partial graph $G_t$, carrying on with the \ac{FLAGG} loop.
%Note that plane features $\pi$ are fixed inputs because they are defined only for the observed planes.
\ac{MiDi}~\citep{vignac2023midi}, a diffusion model capable of sampling both topology and 3D geometry of a graph, is our Fill module of choice. Because it works on the dense representation of a graph, made of tensors, we introduce the following objects, retrievable from the features above: the node labels $\fX_t\in\{0,1\}^{m_t\times|\concepts|}$, the new-new and new-old edge tensor $\fE_t\in\{0,1\}^{m_t\times(m_t+n_{t-1})\times2}$, and the node centroids $\fR_t\in\R^{m_t\times3}$. Finally, let the tensor subgraph $\fZ_t=(\fX_t,\fE_t,\fR_t)$ collect all of these tensors. $\fZ_t$ only considers new nodes and connections because \ac{FLAGG} stores and processes the partial graph $G_{t-1}$ as a sparse graph, leading to lower memory usage. 
Using \ac{MiDi} as a Fill module requires some adaptations because the generative objectives and inputs differ from those of the original work. In the following we detail how MiDi works in this adapted setup at every iteration $t$ of \ac{FLAGG}. First, the coordinate system is anchored to the centroid $\bm c_{t-1}=\frac{1}{n_{t-1}}\sum_{v\in\verts_{t-1}}\rho(v)$ of $G_{t-1}$
and its coordinates are translated as $\bar\rho_{t-1}(v)=\rho_{t-1}(v)-\bm c_{t-1}$, obtaining the centered conditioning graph $\bar G_{t-1}$.
Second, let $s=0,\dots,S$ denote the diffusion time step, where $\fZ_t^0$ is the clean subgraph while $\fZ_t^S$ is the fully noisy subgraph. Then, the MiDi module is trained to reverse the multimodal forward diffusion process at each timestep $s$ given the partial graph $G_{t-1}$. At generation time, MiDi starts from a noisy sample $\fZ_t^S$, and gradually denoises it until it recovers the learned output distribution, sampling $\fZ_t^0$. Refer to Appendix~\ref{sec:appendix_midi} for more details.
The neural network of the adapted MiDi is implemented by a Graph Transformer~\citep{dwivedi2020generalization} with its layers being interleaved with GINE layers processing the nodes of $G_{t-1}$, which are used as conditioning information. At the end of the Fill step, after the graph $G_t$ is obtained, its coordinates are translated back to the original frame by adding $\bm c_{t-1}$. 
FLAGG with MiDi as its Fill module is equivariant in distribution under arbitrary translations and rotations about the vertical axis. The formal statement and proof are given in Theorem~\ref{thm:flagg_equivariance} in Appendix~\ref{sec:appendix_midi_equivariance}.

\subsection{Additional training techniques}
\label{sec:generation_training_techniques}
A potential issue carried by autoregressive architectures is exposure bias~\citep{ranzato2015sequence}\add{, where errors from earlier steps accumulate during generation}. While the literature offers a plethora of solutions to this problem~\citep{schmidt2019generalization,huang2026self}, adapting them to graphs is not trivial. To mitigate this issue, we propose two noise data augmentation schemes \add{that teach the model to recover from imperfect choices:} the first is applied to the Fill module, adding Gaussian noise to node coordinates and random perturbations in node labels and edges; the second is applied to the whole pipeline, diverging from the classical optimal categorical removal sequence used in \ac{FLAGG}, and allowing suboptimal choices, mimicking wrong predictions made by the model.
On the other hand, \ac{FLAGG} is a modular framework, requiring a balance in gradient contributions from each model, and in the diffusion-based Fill module itself. For this reason, we reweight the sample rate of timesteps $t$ in the removal process to favor times with higher variability of $G_t$, allowing the model to explore partial graphs more uniformly, and avoid overfitting to trivial cases at $t=0$ and $t=T$. Focusing on the Fill module, we apply a loss reweighting scheme to the coordinate term in Eq.~\ref{eq:midi_loss} based on SNR~\citep{hang2023snr}, increasing the contribution of almost-clean, low $s$ timesteps and decreasing high noise contributions of high $s$ timesteps. More details can be found in Appendix~\ref{sec:appendix_training_techniques}.

\section{Hierarchical 3DSG generation benchmark}\label{sec:bench}

\subsection{Datasets}\label{sec:bench/data}
To the best of our knowledge, no public dataset provides the complete metric-semantic hierarchical \acp{3DSG} targeted in this work. Therefore, we construct and publicly release our dataset\footnote{\url{https://anonymous.4open.science/r/3DSG-generation-datasets-2787}} consisting of 9 subsets derived from three sources (Tab.~\ref{tab:datasets}):
(i) a synthetic generator (S), which allows controlled layout complexity and for which we define graph structure and node centroids by construction;
(ii) the publicly available MSD dataset (M)~\citep{msd}, which provides structure and centroids for walls and rooms, while higher-level nodes (floor and above) are constructed; and
(iii) real environments (R) recorded by a robot with a LiDAR sensor, where only planes are observed, and thus the full hierarchy (walls, rooms, and higher levels) is manually constructed.
Whenever structure and/or centroids are missing in the source, we infer them using the same node labeling and centroid definition functions, $\lambda(\cdot)$ and $\rho(\cdot)$, ensuring a consistent construction procedure across all sources. For each source, we instantiate subsets at increasing hierarchy depth: up to the floor level (-F), the building level (-B), or the city level (-C). The S-F-$\square$ variant further restricts synthetic layouts to rectangular rooms only.
Layout complexity and graph size grow with hierarchy depth across all sources, while the average node degree remains approximately constant due to the shared tree-like structure. The MSD and real sources exhibit substantially larger and more complex graphs than the synthetic source. The real source is small and used only for testing.

% https://anonymous.4open.science/r/3DSG-generation-datasets-2787
% https://github.com/JoseAndresMR/3DSG-generation-datasets

\subsection{Metrics}\label{sec:bench/metrics}
Generated \acp{3DSG} are compared with ground truth, striving for similar topological statistics and correctness. Thus, we define a set of \textit{graph matching metrics}, comparing the sampled extension of a set of planes with its ground truth, and a set of \textit{distribution metrics}, computing the similarity in distribution of topological properties across the two sets of data.

\textbf{Graph matching evaluation.}
% Given a set of planes, the output of the model must align with the ground truth hierarchy and centroids. Such an alignment can be computed by mapping each generated node to a ground truth node, which is not trivial when the number of nodes mismatches, or the structures are fundamentally different. A mapping for graphs with potentially different number of nodes is obtained through the Fused Gromov-Wasserstein (FGW) distance~\cite{vayer2020fgw}. FGW finds a transport plan between the generated nodes and test nodes, minimizing the transport cost, and can be efficiently computed using the POT library~\citep{flamary2021pot}.
% The cost combines the Wasserstein distance (W), matching nodes by their features, and the Gromov-Wasserstein distance (GW), accounting for the structure:
\add{The output of the model must align with the ground truth hierarchy and centroids, which requires mapping each generated node to a ground truth node even when the number of nodes or the structures differ. We obtain this mapping through the Fused Gromov-Wasserstein (FGW) distance~\citep{vayer2020fgw}, which finds the transport plan between generated and ground truth nodes minimizing a cost that combines the Wasserstein distance (W), matching nodes by their features, and the Gromov-Wasserstein distance (GW), accounting for the structure:}

\begin{equation}
    d_{FGW,\alpha}(G,G') =
    \inf_{\gamma\in\Gamma}(1-\alpha)d_W(\gamma,G,G')+\alpha d_{GW}(\gamma,G,G'),
\end{equation}
where $\gamma:\verts\times\verts'\to[0,1]$ is a transport plan in the set of all plans $\Gamma$, and $\alpha=0.5$.
The W distance matches nodes by their centroids and one-hot encoded labels, while the GW distance matches pairs of nodes by their shortest-path lengths $\operatorname{SP}$ in $G$ and $G'$, as a measurement of graph topology similarity:
\begin{gather}
    d_W(\gamma,G,G')=\sum_{(v,v')\in\verts\times\verts'}\gamma(v,v')\bigg\lVert\begin{pmatrix}
        \rho(v)\\\operatorname{onehot}(\lambda(v))
    \end{pmatrix}-\begin{pmatrix}
        \rho'(v')\\\operatorname{onehot}(\lambda'(v'))
    \end{pmatrix}\bigg\rVert,\\
        d_{GW}(\gamma,G,G')=
        \sum_{\substack{(v,w)\in\verts\times\verts\\(v',w')\in\verts'\times\verts'}}\gamma(v,v')\gamma(w,w') \|\operatorname{SP}(v,w)-\operatorname{SP}(v',w')\|.
\end{gather}
%To allow a 1-to-1 matching between the two graphs, we add the difference in nodes to the smallest one.
To allow a 1-to-1 matching between the two graphs, we pad the smallest with $|n - n'|$ isolated dummy nodes whose pairwise $\operatorname{SP}$ lengths to all other nodes are set to one plus the maximum $\operatorname{SP}$ in the original graph. This ensures that dummy nodes are penalized by both the W and GW terms, so the transport plan avoids matching real nodes to dummy nodes unless a better alignment exists.

\textbf{Distribution evaluation.}
Evaluating distribution similarity is non-trivial in graphs, as they present a broad range of properties~\citep{thompson2022evaluation}. Current metrics are based on the Maximum Mean Discrepancy (MMD) comparing the distance between the distributions of a feature. Please find details in Appendix~\ref{sec:app_metrics}. As features we consider: the node degree histogram (Deg.), the clustering coefficient histogram (Clust.), the spectrum (Spec.), and the embeddings obtained by a randomly initialized Graph Isomorphism Network (GIN)~\citep{xu2018gin}.

\begin{table}[!t]
\centering
\caption{\textbf{Datasets.} Each level includes all the levels below it: floor-level graphs contain walls, rooms and floors, building-level graphs add buildings, and city-level graphs add cities. Per-graph averages.}
\label{tab:datasets}
\setlength{\arraycolsep}{2pt}
\renewcommand{\arraystretch}{1}
\resizebox{\columnwidth}{!}{%
\begin{tabular}{|c|ccc|ccc|ccc|}
\hline
\textbf{Source} &
\multicolumn{3}{c|}{\textbf{Synthetic}} &
\multicolumn{3}{c|}{\textbf{MSD}} &
\multicolumn{3}{c|}{\textbf{Real}} \\ \hline

\textbf{Highest level} &
Floor & Floor & City &
Floor & Building & City &
Floor & Building & City \\
\textbf{Dataset} &
S-F-$\square$ ($\square$ rooms) & S-F & S-C &
M-F & M-B & M-C &
R-F & R-B & R-C \\ \hline

\textbf{\# graphs}    & 2{,}400 & 2{,}999 & 2{,}999 & 3{,}466 & 2{,}416 & 1{,}084 & 4 & 1 & 1 \\
{\boldmath$|\verts|$}     & 23   & 30   & 110  & 95   & 159  & 223  & 45   & 135  & 172  \\
{\boldmath$|\edges|$}     & 26.2 & 34.8 & 130.5 & 112.3 & 186.8 & 258.6 & 46.0 & 145.0 & 185.0 \\
\textbf{Degree}           & 2.28 & 2.34 & 2.38 & 2.36 & 2.35 & 2.32 & 2.04 & 2.13 & 2.11 \\ \hline
\end{tabular}
}
\end{table}

\section{Experiments}
\label{sec:experiments}

We evaluate all models and baselines on the proposed 3DSG benchmark (see Section~\ref{sec:bench}). Specifically, we split each dataset in Section~\ref{sec:bench/data} into 60\% for training, 20\% for validation and 20\% for test. We train our models on the training set, and select the checkpoint scoring the best FGW on the validation set. Finally, we report the metrics in Section~\ref{sec:bench/metrics} computed on the test set in Table~\ref{tab:stacked_three_sets} and Table~\ref{tab:real_exp}. All hyperparameters can be found in Appendix~\ref{sec:appendix_hyperparameters}. \add{We run all experiments with three seeds and report the mean and standard deviation across them. Experiments were run on NVIDIA V100-32GB GPUs.} We make our code publicly available\footnote{\url{https://anonymous.4open.science/r/3dsg-gen-iclr2027}}.

% https://anonymous.4open.science/r/3dsg-gen-iclr2027

\subsection{Baselines}

\add{We compare with two learning-based methods for this task, EC-$\square$ and EC-L (Sec.~\ref{sec:related_works}), and a one-shot diffusion model; Tab.~\ref{tab:methods} summarizes their capabilities.
We exclude geometry-driven pipelines that operate on raw sensor maps rather than on graphs of planes.
EC-$\square$ (rectangular rooms) and EC-L (L-shaped rooms) are trained on their original data (S-F-$\square$
and S-F, respectively) and, since they generate no floors, are evaluated on S-F-$\square$, S-F, M-F and R-F with floor nodes removed, over three
random seeds.} %To enable a comparison across hierarchical levels \add{to a lower-bound reference}, we also identify a random-guessing baseline called \textsc{Rand}. \textsc{Rand} is run with three different random seeds on every dataset on which it is evaluated, and we report results on the test split. More details are found in Appendix~\ref{sec:appendix/base/random}.}
\add{Finally, to compare against the state of the art in graph generation, we include \textsc{MiDi+Or}, the one-shot model \ac{MiDi}~\citep{vignac2023midi} adapted to conditional 3DSG completion as described in Sec.~\ref{subsec:one_shot_gen}. Since one-shot models require the final graph size as input, \textsc{MiDi+Or} receives the correct number of nodes to be generated from an oracle (Or), an advantage not available to autoregressive methods. \textsc{MiDi+Or} generates the full hierarchy and is therefore evaluated on all datasets. We regard \textsc{MiDi+Or} as a reference under privileged information rather than a directly deployable method. Since \textsc{MiDi+Or} relies on the same denoising model, the comparison between the two highlights the effect of the generation scheme, autoregressive versus one-shot. We provide an additional random baseline as upper-bound reference in Appendix.~\ref{sec:appendix/base/random}}

\subsection{Results}
\label{sec:experiments_results}

\begin{figure*}[t!]
  \begin{center}
    \centerline{\includegraphics[width=\textwidth]{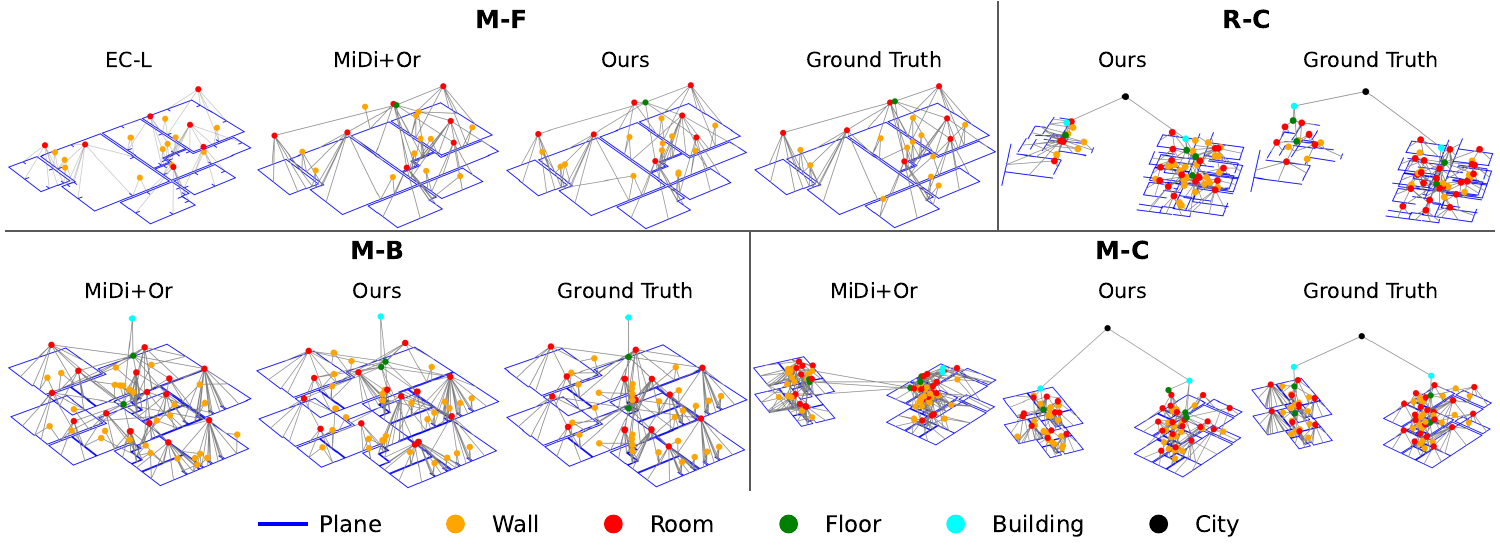}}
    \caption{
      \add{\textbf{Qualitative results.} Selection of the most complex datasets at the three hierarchy levels explored, comparing our model with the best-performing baselines. Full comparison  in the Appendix~\ref{sec:appendix_qualitative_results}.}
    }
    \label{fig:qualitative_results}
  \end{center}
\end{figure*}

Tab.~\ref{tab:stacked_three_sets} and Fig.~\ref{fig:qualitative_results} report the results on the synthetic and MSD sources. \textsc{Ours} attains the lowest FGW on every dataset but M-F. Against EC-$\square$ and \textsc{EC-L}, the gain is large even on the single-floor datasets on which they were trained (0.8--1.1 vs.\ 8.0--13.0 in FGW) and grows with layout complexity: on M-F, FGW, GW and W drop by 78--80\% relative to \textsc{EC-L}, i.e., structure and metric-semantic attributes improve alike. The distribution metrics agree: on M-F, Degree and Spec.\ drop by 93\% and 98\%, and GIN from 6.0 to 2.3. On clustering, \textsc{Ours} matches the zero error that EC-$\square$ and \textsc{EC-L} obtain by construction, while learning the full adjacency.
More importantly, \textsc{Ours} surpasses \textsc{MiDi+Or}, which shares the same denoising model but receives the target graph size from an oracle, on every dataset but M-F, and the gap widens with depth (S-C: 3.0 vs.\ 6.0; M-B: 4.8 vs.\ 9.3; M-C: 7.3 vs.\ 13.3). The advantage of \textsc{MiDi+Or} lies mainly in GW on single-floor datasets, where it scores close to zero, as expected when the exact number of nodes is given, and yields a lower FGW on M-F (3.5 vs.\ 4.3). Beyond one floor, however, its errors grow and become unstable across seeds (M-C: FGW $13.3 \pm 13.6$, Clust.\ $72.2 \pm 42.2$), whereas \textsc{Ours} degrades gracefully (FGW 4.3, 4.8, 7.3 on M-F, M-B, M-C) with a standard deviation never above 0.7. Further results on memory and generation time can be found in Appendix~\ref{sec:appendix_quantitative_results}.
% As the Tab.~\ref{tab:ablation} presents, both noisy data augmentation schemes are fundamental for high quality sampling, confirming the hypothesis of exposure bias without them. \textsc{Ours-FD4 filler} shows promising performance in on the structural part of FGW (FW), but presents worse distribution metrics, in line with worse W metric, suggesting that it struggles capturing the distances and, hence, positions.

On robot-recorded environments (Tab.~\ref{tab:real_exp}), evaluated with the models trained on the MSD subset of matching depth, all learned methods lose accuracy, but to very different extents. \textsc{Ours} roughly triples its floor-level FGW (4.3 on M-F to 13.2 on R-F), while R-B and R-C degrade less (8.5 and 10.1). \textsc{MiDi+Or} loses most: on R-F its FGW rises from 3.5 to 33.7, falling behind \textsc{EC-L}, with the structural term reaching GW 53.3, and on R-B it becomes unstable across seeds ($25.4 \pm 21.8$). \textsc{Ours} thus attains the lowest FGW on every subset, 40\% below \textsc{EC-L} on R-F and 67--74\% below \textsc{MiDi+Or} on R-B and R-C, and the oracle keeps an advantage only in W on R-F. Fig.~\ref{fig:qualitative_results} and Appendix~\ref{sec:appendix_qualitative_results} show the failure mode: \textsc{MiDi+Or} attaches rooms and floors to planes across the whole scene, whereas \textsc{Ours} preserves the local grouping of the ground truth.

\begin{table*}[t]
\centering
\small
\setlength{\tabcolsep}{3.5pt}
\caption{\textbf{Evaluation results for synthetic and MSD datasets.} Values are multiplied by $10^2$ and reported as mean $\pm$ standard deviation over three seeds. Missing values (-) occur when the dataset contains hierarchy levels out of the scope of the baseline. Best results are highlighted in light gray and bold; second-best results in lighter gray.}
\label{tab:stacked_three_sets}
\resizebox{\textwidth}{!}{%
\begin{tabular}{l ccccccc ccccccc ccccccc}
\toprule
 
% ===================== SET 1 =====================
\multicolumn{22}{l}{\textbf{Source: Synthetic}} \\
\midrule
\multirow{2}{*}{Method} &
\multicolumn{7}{c}{S-F-$\square$} &
\multicolumn{7}{c}{S-F} &
\multicolumn{7}{c}{S-C} \\
\cmidrule(lr){2-8}\cmidrule(lr){9-15}\cmidrule(lr){16-22}
& FGW $\downarrow$ & GW$\downarrow$ & W$\downarrow$ & Deg.$\downarrow$ & Spec.$\downarrow$ & Clust.$\downarrow$ & GIN$\downarrow$
& FGW $\downarrow$ & GW$\downarrow$ & W$\downarrow$ & Deg.$\downarrow$ & Spec.$\downarrow$ & Clust.$\downarrow$ & GIN$\downarrow$
& FGW $\downarrow$ & GW$\downarrow$ & W$\downarrow$ & Deg.$\downarrow$ & Spec.$\downarrow$ & Clust.$\downarrow$ & GIN$\downarrow$ \\
\midrule
\textsc{EC-$\square$} &
\nlvalpm{12.0}{1.0} & \nlvalpm{17.0}{1.0} & \nlvalpm{8.0}{0.0} & \nlvalpm{18.0}{1.0} & \nlvalpm{12.0}{1.0} & \best{\nlvalpm{0.0}{0.0}} & \nlvalpm{40.0}{10.0} &
\nlvalpm{13.0}{1.0} & \nlvalpm{17.0}{1.0} & \nlvalpm{8.0}{1.0} & \second{\nlvalpm{19.0}{2.0}} & \nlvalpm{12.0}{1.0} & \best{\nlvalpm{0.0}{0.0}} & \nlvalpm{36.0}{9.0} &
- & - & - & - & - & - & - \\
\textsc{EC-L} &
\nlvalpm{8.0}{0.0} & \nlvalpm{7.0}{0.0} & \nlvalpm{9.0}{0.0} & \nlvalpm{25.0}{2.0} & \nlvalpm{8.0}{1.0} & \best{\nlvalpm{0.0}{0.0}} & \nlvalpm{16.0}{1.0} &
\nlvalpm{8.0}{0.0} & \nlvalpm{7.0}{0.0} & \nlvalpm{9.0}{0.0} & \nlvalpm{26.0}{2.0} & \nlvalpm{8.0}{1.0} & \best{\nlvalpm{0.0}{0.0}} & \second{\nlvalpm{20.0}{2.0}} &
- & - & - & - & - & - & - \\
\midrule
\textsc{\add{MiDi+Or}} &
\second{\nlvalpm{3.4}{0.1}} & \best{\nlvalpm{0.1}{0.2}} & \second{\nlvalpm{6.7}{0.1}} & \best{\nlvalpm{0.2}{0.1}} & \best{\nlvalpm{0.5}{0.1}} & \best{\nlvalpm{0.0}{0.0}} & \best{\nlvalpm{0.2}{0.0}} &
\second{\nlvalpm{3.3}{0.0}} & \best{\nlvalpm{0.0}{0.0}} & \second{\nlvalpm{6.5}{0.0}} & \best{\nlvalpm{0.4}{0.1}} & \best{\nlvalpm{0.5}{0.1}} & \best{\nlvalpm{0.0}{0.0}} & \best{\nlvalpm{0.4}{0.0}} &
\second{\nlvalpm{6.0}{0.3}} & \second{\nlvalpm{7.3}{1.3}} & \second{\nlvalpm{4.8}{0.8}} & \second{\nlvalpm{33.3}{25.8}} & \second{\nlvalpm{6.8}{1.8}} & \second{\nlvalpm{1.4}{1.0}} & \second{\nlvalpm{12.7}{10.5}} \\
\midrule
\textsc{\add{Ours}} &
\best{\nlvalpm{0.8}{0.1}} & \second{\nlvalpm{0.7}{0.1}} & \best{\nlvalpm{1.0}{0.1}} & \second{\nlvalpm{0.3}{0.1}} & \second{\nlvalpm{1.0}{0.1}} & \best{\nlvalpm{0.0}{0.0}} & \second{\nlvalpm{0.3}{0.0}} &
\best{\nlvalpm{1.1}{0.1}} & \second{\nlvalpm{1.0}{0.1}} & \best{\nlvalpm{1.3}{0.1}} & \best{\nlvalpm{0.4}{0.0}} & \second{\nlvalpm{0.6}{0.0}} & \best{\nlvalpm{0.0}{0.0}} & \best{\nlvalpm{0.4}{0.0}} &
\best{\nlvalpm{3.0}{0.5}} & \best{\nlvalpm{3.2}{0.8}} & \best{\nlvalpm{2.8}{0.3}} & \best{\nlvalpm{2.8}{0.8}} & \best{\nlvalpm{1.6}{0.1}} & \best{\nlvalpm{0.1}{0.2}} & \best{\nlvalpm{2.0}{1.1}} \\
\midrule
 
% ===================== SET 2 =====================
\multicolumn{22}{l}{\textbf{Source: MSD}} \\
\midrule
\multirow{2}{*}{Method} &
\multicolumn{7}{c}{M-F} &
\multicolumn{7}{c}{M-B} &
\multicolumn{7}{c}{M-C} \\
\cmidrule(lr){2-8}\cmidrule(lr){9-15}\cmidrule(lr){16-22}
& FGW $\downarrow$ & GW$\downarrow$ & W$\downarrow$ & Deg.$\downarrow$ & Spec.$\downarrow$ & Clust.$\downarrow$ & GIN$\downarrow$
& FGW $\downarrow$ & GW$\downarrow$ & W$\downarrow$ & Deg.$\downarrow$ & Spec.$\downarrow$ & Clust.$\downarrow$ & GIN$\downarrow$
& FGW $\downarrow$ & GW$\downarrow$ & W$\downarrow$ & Deg.$\downarrow$ & Spec.$\downarrow$ & Clust.$\downarrow$ & GIN$\downarrow$ \\
\midrule
\textsc{EC-$\square$} &
\nlvalpm{27.0}{0.0} & \nlvalpm{27.0}{0.0} & \nlvalpm{28.0}{0.0} & \nlvalpm{80.0}{0.0} & \nlvalpm{116.0}{0.0} & \best{\nlvalpm{0.0}{0.0}} & \nlvalpm{100.0}{1.0} &
- & - & - & - & - & - & - &
- & - & - & - & - & - & - \\
\textsc{EC-L} &
\nlvalpm{20.0}{0.0} & \nlvalpm{18.0}{0.0} & \nlvalpm{23.0}{0.0} & \nlvalpm{31.0}{0.0} & \nlvalpm{52.0}{0.0} & \best{\nlvalpm{0.0}{0.0}} & \nlvalpm{6.0}{0.0} &
- & - & - & - & - & - & - &
- & - & - & - & - & - & - \\
\midrule
\textsc{\add{MiDi+Or}} &
\best{\nlvalpm{3.5}{0.1}} & \best{\nlvalpm{0.2}{0.0}} & \second{\nlvalpm{6.9}{0.2}} & \second{\nlvalpm{3.3}{1.7}} & \second{\nlvalpm{1.4}{0.2}} & \best{\nlvalpm{0.0}{0.0}} & \best{\nlvalpm{1.2}{0.3}} &
\second{\nlvalpm{9.3}{8.1}} & \second{\nlvalpm{9.7}{14.0}} & \second{\nlvalpm{8.9}{2.2}} & \second{\nlvalpm{46.8}{42.8}} & \second{\nlvalpm{15.0}{15.0}} & \second{\nlvalpm{28.8}{42.5}} & \second{\nlvalpm{15.8}{16.6}} &
\second{\nlvalpm{13.3}{13.6}} & \second{\nlvalpm{20.7}{26.9}} & \best{\nlvalpm{6.0}{0.5}} & \second{\nlvalpm{59.8}{17.7}} & \second{\nlvalpm{36.1}{40.9}} & \second{\nlvalpm{72.2}{42.2}} & \second{\nlvalpm{41.7}{24.7}} \\
\midrule
\textsc{\add{Ours}} &
\second{\nlvalpm{4.3}{0.0}} & \second{\nlvalpm{3.9}{0.1}} & \best{\nlvalpm{4.6}{0.1}} & \best{\nlvalpm{2.2}{0.4}} & \best{\nlvalpm{1.0}{0.1}} & \best{\nlvalpm{0.0}{0.0}} & \second{\nlvalpm{2.3}{0.5}} &
\best{\nlvalpm{4.8}{0.1}} & \best{\nlvalpm{4.5}{0.1}} & \best{\nlvalpm{5.1}{0.1}} & \best{\nlvalpm{5.4}{1.7}} & \best{\nlvalpm{2.9}{0.5}} & \best{\nlvalpm{0.0}{0.0}} & \best{\nlvalpm{4.2}{2.4}} &
\best{\nlvalpm{7.3}{0.7}} & \best{\nlvalpm{7.0}{0.4}} & \second{\nlvalpm{7.6}{1.1}} & \best{\nlvalpm{11.4}{3.1}} & \best{\nlvalpm{8.4}{3.3}} & \best{\nlvalpm{0.2}{0.2}} & \best{\nlvalpm{21.0}{19.0}} \\
\bottomrule
\end{tabular}
}
\end{table*}

\begin{table*}[t]
\centering
\footnotesize
% ---------- Right: real environments (10 columns) ----------
\begin{minipage}[t]{0.46\textwidth}
\centering
\caption{\textbf{Results in real environments.}} %Values are multiplied by $10^2$ and reported as mean $\pm$ standard deviation.}
\label{tab:real_exp}
\footnotesize
\setlength{\tabcolsep}{2.5pt}
\begin{adjustbox}{max width=\columnwidth}
\begin{tabular}{l ccc ccc ccc}
\toprule
\multicolumn{10}{l}{\textbf{Source: Real [$\times 10^{-2}$]}} \\
\midrule
\multirow{2}{*}{Method} &
\multicolumn{3}{c}{R-F} &
\multicolumn{3}{c}{R-B} &
\multicolumn{3}{c}{R-C} \\
\cmidrule(lr){2-4}\cmidrule(lr){5-7}\cmidrule(lr){8-10}
& FGW $\downarrow$ & GW$\downarrow$ & W$\downarrow$
& FGW $\downarrow$ & GW$\downarrow$ & W$\downarrow$
& FGW $\downarrow$ & GW$\downarrow$ & W$\downarrow$ \\
\midrule
\textsc{EC-$\square$} &
\nlvalpm{30.4}{0.0} & \nlvalpm{38.0}{0.0} & \nlvalpm{22.9}{0.0}
& - & - & -
& - & - & - \\
\textsc{EC-L} &
\second{\nlvalpm{22.0}{0.0}} & \second{\nlvalpm{27.0}{0.0}} & \nlvalpm{17.0}{0.0}
& - & - & -
& - & - & - \\
\midrule
\textsc{MiDi+Or} &
\nlvalpm{33.7}{5.7} & \nlvalpm{53.3}{10.1} & \best{\nlvalpm{14.1}{2.9}} &
\second{\nlvalpm{25.4}{21.8}} & \second{\nlvalpm{36.8}{40.5}} & \second{\nlvalpm{14.0}{6.4}} &
\second{\nlvalpm{38.1}{9.8}} & \second{\nlvalpm{61.7}{22.4}} & \second{\nlvalpm{14.5}{3.3}} \\
\textsc{Ours} &
\best{\nlvalpm{13.2}{1.3}} & \best{\nlvalpm{9.9}{0.9}} & \second{\nlvalpm{16.5}{1.8}} &
\best{\nlvalpm{8.5}{3.6}} & \best{\nlvalpm{6.5}{4.5}} & \best{\nlvalpm{10.5}{3.4}} &
\best{\nlvalpm{10.1}{5.4}} & \best{\nlvalpm{10.8}{6.9}} & \best{\nlvalpm{9.4}{4.4}} \\
\bottomrule
\end{tabular}
\end{adjustbox}
\end{minipage}\hfill
% ---------- Left: ablation ----------
\begin{minipage}[t]{0.53\textwidth}
\centering
\caption{\textbf{Ablation study in S-C}. At 40k training steps.} %Every variant is trained for 40k steps and reported as mean $\pm$ standard deviation over two seeds.}
\label{tab:ablation}
\footnotesize
\setlength{\tabcolsep}{3.5pt}
\resizebox{\columnwidth}{!}{%
\begin{tabular}{l ccccccc}
\toprule
\multirow{2}{*}{Method} &
\multicolumn{7}{c}{S-C [$\times 10^{-2}$]} \\
\cmidrule(lr){2-8}
& FGW $\downarrow$ & GW$\downarrow$ & W$\downarrow$ & Deg.$\downarrow$ & Spec.$\downarrow$ & Clust.$\downarrow$ & GIN$\downarrow$ \\

\midrule
w/o noisy blocks &
6.89 & 7.68 & 6.09 & \second{25.52} & \second{5.71} & \best{0.00} & 4.35 \\
w/o noise aug. &
12.28 & 11.28 & 13.28 & 44.33 & 16.11 & 1.24 & \second{3.75} \\
\midrule
Ours-FD4 filler &
\best{5.78} & \best{5.76} & \second{5.80} & 27.84 & 7.18 & 0.41 & 11.64 \\
\midrule
Ours &
\second{6.10} & \second{6.86} & \best{5.34} & \best{18.73} & \best{4.89} & \second{0.05} & \best{3.11} \\
\bottomrule
\end{tabular}
}
\end{minipage}
\end{table*}

\subsection{Ablations}
We ablate the two exposure bias mitigation strategies of Sec.~\ref{sec:generation_training_techniques} on the validation set of S-C, which combines non-rectangular layouts with full hierarchy depth. The variants in this study are trained for 40k steps, differently from other experiments which are trained to convergence. Two leave-one-out variants remove either the perturbation of the removal sequence (\emph{w/o noisy blocks}) or that of the conditioning graph (\emph{w/o noise aug.}). A third variant keeps both perturbations but replaces MiDi with FD4 in the Fill module (\emph{Ours-FD4 filler}), isolating how block geometry is represented: instead of denoising the 3D coordinates of new nodes, FD4 denoises edge distances and recovers centroids via Multidimensional Scaling (Appendix~\ref{sec:appendix_fd4}).
% Preambl

As Tab.~\ref{tab:ablation} shows, both noise schemes improve sampling, consistent with exposure bias: without noise augmentation FGW doubles (12.28 vs.\ 6.10) and every metric worsens, while without noisy blocks FGW rises to 6.89, mainly through degree and GIN. \textsc{Ours-FD4 filler} reaches the lowest FGW (5.78) through a lower GW, but a higher W and clearly worse distribution metrics (GIN 11.64 vs.\ 3.11), suggesting that recovering centroids from denoised distances is less accurate and that its per-graph structural matches do not yield realistic graph statistics.
\section{Discussion and limitations}
\label{sec:discussion}

Three conclusions follow from Sec.~\ref{sec:experiments}. First, jointly learning structure, labels and centroids in one model outperforms the edge-classification pipelines, most on the layouts that break their assumptions (concepts composed of a variable number of lower-level nodes), and removes their ceiling: EC-$\square$ and \textsc{EC-L} cannot represent anything above rooms, and their zero clustering error only reflects that they emit two edge types, whereas \textsc{Ours} learns the full adjacency and still attains it.
Second, the autoregressive scheme improves generation itself, beyond enabling variable graph sizes: with the same denoising model and the ground-truth number of nodes, one-shot generation is competitive only on single floors and falls behind, with growing variance, as the hierarchy deepens. Since every concept in a 3DSG is defined by the lower-level nodes it aggregates, generating one block at a time conditions each new node on the layer that supports it, whereas one-shot generation must resolve all layers and their dependencies at once.
Third, the remaining error is driven mainly by layout complexity: FGW quadruples from synthetic to MSD single-floor scenes but rises only mildly when floors, buildings and cities are added, so the same architecture, trained per hierarchy level without concept-specific modules, extends to deeper hierarchies at a moderate cost. The largest source of error is the shift from floor plans to sensor data, which affects every method; that the one-shot model loses most of its accuracy there while \textsc{Ours} retains most of it suggests that conditioning on the layer below makes generation less sensitive to the input geometry.
%Finally, GW tracks the degree-, spectral-, clustering- and GIN-based scores, making it a compact proxy for structural quality; with W for metric-semantic fidelity, FGW is a well-balanced single indicator for 3DSG generation.
Finally, the proposed FGW-based metrics proved useful for assessing topological and spatial similarity with respect to ground truth hierarchies. They are complementary to standard graph generation indicators, offering 3DSG-specific measures of quality: GW for the structure, W for metric-semantic fidelity, FGW as a well-balanced single indicator for 3DSG generation.

\section{Conclusion}

We propose the first autoregressive diffusion-based \ac{DGG} model for \ac{3DSG} completion, which grows a hierarchy of walls, rooms, floors, buildings and cities from robot-observed planes, jointly generating structure, labels and centroids without concept-specific modules. Across synthetic scenes, architectural floor plans, and robotic sensor data, it outperforms existing learning-based baselines on single-floor generation, extends to multi-floor, multi-building and city scenarios where no prior learning-based method applies, and, without access to the target graph size, surpasses a one-shot model that receives it from an oracle on all but one dataset, with the gap widening as the hierarchy deepens. We further advocate Fused Gromov--Wasserstein (FGW) as a unified metric for \ac{3DSG} generation and publicly release our dataset and code.
Two limitations point to future work. The shift from floor plans to robot-recorded planes remains the largest source of error for all methods, \textsc{Ours} included; since real data are used only for testing, training on or adapting to sensor-derived planes is a natural next step. Moreover, our setting is offline and bottom-up, so an early error propagates upward through the hierarchy; online generation that refines the hierarchy as new planes arrive would address both and enable use during mapping.

\section*{Acknowledgements}
This work was funded, in whole or in part, by the Fonds National de la
Recherche of Luxembourg (FNR), DEUS Project (Ref.\ C22/IS/17387634/DEUS)
and RoboSAUR Project (Ref.\ 17097684/RoboSAUR); the PNRR project FAIR --- Future AI Research (PE00000013), under the NRRP MUR program funded by the NextGenerationEU; the Italian Ministry of University and Research (MUR) under the PRIN program – Progetti di Rilevante Interesse Nazionale – PRIN 2022 (Secretary-General's Decree No. 1401 of 18/09/2024) – CUP C53C24000770006, project title ``DEEP-GRAPH: Design and Theory of Deep Graph Learning''.

We thank Dr.\ Cecilia Morales and Dr.\ Miguel Fernandez-Cortizas for their valuable feedback and help in improving the writing of this manuscript.

\subsection*{AI use statement}

% (This section is \textbf{required} and does not count toward the page limit.)

In this work, we used generative AI tools to aid or polish writing, research ideation or execution, and for retrieval and discovery (e.g., finding related work).
We have not used generative AI tools for drafting sections of the paper, generating synthetic datasets, or proving mathematical claims.
Additionally, we used generative AI tools to revise our own code for errors. We have reviewed all AI-assisted work. We checked LLM-generated research ideas for potential plagiarism through a manual literature survey and LLM-generated code was verified and tested for correctness by the two first authors. We take responsibility for the final content of this work,
including text, claims or artifacts produced with the aid of generative AI.

\subsection*{Reproducibility statement}
We have taken several steps to make our results reproducible. The complete source code, including training, sampling, and evaluation scripts, is available at an anonymous repository (Sec.~\ref{sec:experiments}), and the hierarchical 3DSG datasets used in all experiments, covering the synthetic, MSD-derived, and robot-recorded subsets, are publicly released (Sec.~\ref{sec:bench}). Sec.~\ref{sec:bench/data} and Tab.~\ref{tab:datasets} describe how each subset is constructed, including how missing structure and centroids are completed with a consistent labeling and centroid definition across sources. The model is specified in Section~\ref{sec:generating_3dsg}, with the node orderings in Appendix~\ref{sec:method/order}, the training techniques in Appendix~\ref{sec:appendix_training_techniques}, the FD4 variant in Appendix~\ref{sec:appendix_fd4}, and the baselines in Appendix~\ref{subsec:one_shot_gen}. Appendix~\ref{sec:appendix_hyperparameters} includes a comprehensive list of relevant hyperparameter of our model and of MiDi+Or, together with the configuration file in the released code that sets each value, so that each run can be reproduced from a single preset. Section~\ref{sec:experiments} specifies the data splits, the checkpoint-selection criterion, the number of seeds, and the hardware. Appendix~\ref{sec:appendix_quantitative_results} reports sampling time and memory, and the metrics are defined in Sec.~\ref{sec:bench/metrics}.

\bibliography{bibliography}
\bibliographystyle{iclr2027_conference}
\newpage

\appendix
\section{MiDi as Fill Module}\label{sec:appendix_midi}
In this section we present the technical details of the MiDi training and generation. The forward diffusion transition is:
\begin{equation}
    q(\fZ_t^s|\fZ_t^{s-1})=\operatorname{Cat}(\fX_t^{s-1} \fQ_X^s)\times \operatorname{Cat}(\fE_t^{s-1} \fQ_E^s)\times\N(\alpha^s\fR_t^{s-1},(\sigma^s)^2\mI),
\end{equation}
where $\fQ_X^s$ and $\fQ_E^s$ are the transition matrices for the discrete diffusion of node labels and edges, respectively, while $\alpha^s$ and $\sigma^s$ are the scaling and noise parameters for the Gaussian diffusion of coordinates. The transitions are built to make the process converge to an easy-to-sample noisy subgraph $\fZ_t^S$. The role of the generative model is to reverse this process starting from $\fZ_t^S$ and the conditioning graph $G_{t-1}$, producing $\fZ_t^0$. The reverse transitions are:
% \begin{align*}
%     p_\theta(\fX_i^{t-1}\mid\gG^t) &= \sum_{a\in\atoms}q(\fX_i^{t-1} \mid \fX_i^{t},\tilde{\rmX}_{ia}^0=1)p_{\theta}^X(\tilde{\rmX}_{ia}^0=1\mid\gG^t) ;\\
%     p_\theta(\fE_{ij}^{t-1}\mid\gG^t) &= \sum_{b\in\bonds}q(\fE_{ij}^{t-1} \mid \fE_{ij}^{t},\tilde{\rmE}_{ijb}^0=1)p_{\theta}^E(\tilde{\rmE}_{ijb}^0=1\mid\gG^t) ;\\
%     p_\theta(\fC_i^{t-1}\mid\gG^t) &= \sum_{c\in\charges}q(\fC_i^{t-1} \mid \fC_i^{t},\tilde{\rmC}_{ic}^0=1)p_{\theta}^C(\tilde{\rmC}_{ic}^0=1\mid\gG^t) ;
% \end{align*}
% \begin{align*}
%     p_\theta(\fD^{t-1}|\fE^{t-1},\gG^t)&=\N_S(\fD^{t-1};\; \mu^t\hat{\fD}_\theta(\gG^t,\fE^{t-1})+\nu^t\fD^t,\, (\tilde{\sigma}^t)^2\mI)  ;\\
%     \mu^t   &=\bar{\alpha}^t(1-(\alpha^t\bar{\sigma}^{t-1} / \bar{\sigma}^t)^2)  ;\\
%     \nu^t   &=\alpha^t (\bar{\sigma}^{t-1} / \bar{\sigma}^t)^2 ;\\
%     \tilde{\sigma}^t    &=(\bar{\sigma}^{t-1})^2(1-(\alpha^t \bar{\sigma}^{t-1} / \bar{\sigma}^t)^2)      ;
% \end{align*}
\begin{equation}
    \label{eq:midi_reverse}
    \begin{aligned}
        p_\theta(\fX_t^{s-1}|\fZ_t^s,\bar G_{t-1}) &= \sum\nolimits_{\fX_t^0} q(\fX_t^{s-1}|\fX_t^s,\fX_t^0)p_\theta^X(\fX_t^0|\fZ_t^s,\bar G_{t-1}) ;\\
        p_\theta(\fE_t^{s-1}|\fZ_t^s,\bar G_{t-1}) &= \sum\nolimits_{\fE_t^0} q(\fE_t^{s-1}|\fE_t^s,\fE_t^0)p_\theta^E(\fE_t^0|\fZ_t^s,\bar G_{t-1}) ;\\
        p_\theta(\fR_t^{s-1}|\fZ_t^s,\bar G_{t-1}) &= \N(\mu^s\hat{\fR}_\theta(\fZ_t^s,\bar G_{t-1})+\nu^s\fR^s_t,\, (\tilde{\sigma}^s)^2\mI)  ;\\
    \end{aligned}
\end{equation}
where $p_\theta(\fX_t^0|\fZ_t^s,\bar G_{t-1})$, $p_\theta(\fE_t^0|\fZ_t^s,\bar G_{t-1})$, and $\hat{\fR}_\theta(\fZ_t^s,\bar G_{t-1})$ are the predicted clean node labels, edges, and coordinates, respectively, while $\mu^s$, $\nu^s$, and $\tilde{\sigma}^s$ are the Gaussian transition parameters. The model, parameterized by $\theta$, is trained to maximize the log-likelihood of the reverse transitions, which is equivalent to minimizing the following loss:
\begin{equation}
    \label{eq:midi_loss}
    \mathcal{L}(\theta) = 
    \lambda_x \CE(\fX_t^0, p_\theta^X(\cdot|\fZ_t^s,\bar G_{t-1})) + \lambda_e \CE(\fE_ t^0,p_\theta^E(\cdot|\fZ_t^s,\bar G_{t-1})) +
    \lambda_r \|\fR_t^0 - \hat{\fR}_\theta(\fZ_t^s,\bar G_{t-1})\|_2^2,
\end{equation}
where $\lambda_x$, $\lambda_e$, and $\lambda_r$ are the weights for the node labels, edges, and coordinates losses, respectively, and $\CE$ is the cross-entropy loss.
\section{Architectural design}

\subsection{Orderings for 3DSGs}\label{sec:method/order}
The ordering $\sigma:\verts\to\{1,\dots,|\verts|\}$ in \ac{FLAGG} is used to define the removal order, which in turn is learned backward. To facilitate understanding, we will refer to $\sigma$ as the intended order of \textit{insertion}, instead of removal. As illustrated in Eq.~\ref{eq:flagg_full}, the probability of a graph $G$ is influenced by the different orderings in which it is generated~\citep{liao2019gran}. As stated in Section~\ref{sec:back/problem}, it is known that the nodes of each layer in the hierarchy directly depend on their children in the layer below. For this reason, a sensible way to order nodes is to do so by layer in a bottom-up fashion. We refer to it as Hierarchical ordering (H). First, an order on the concepts $\sigma_\concepts:\concepts\to\{1,\dots,|\concepts|\}$ must be defined. Then, the ordered layers are defined as
\begin{equation}
    \mathcal{L}_{i} =\left\{v\in\verts: \sigma_\concepts(\lambda(v))=i\right\},\quad\forall i=1,\dots,|\concepts|
\end{equation}
Let $\sigma_{U}(\mathcal{L}):\mathcal{L}\to\{1,\dots,|\mathcal{L}|\}$ be a uniformly random ordering of all the nodes in a generic $\mathcal{L}$. The Hierarchical ordering is given by:
\begin{equation}
    \sigma_\text{H}(v)=\sum_{i=1}^{k-1}|\mathcal{L}_{i}|+\sigma_{U}(\mathcal{L}_{k})(v)\quad\text{with $v\in\mathcal{L}_{k}$}.
\end{equation}
We additionally propose a Hierarchical Proximity ordering (HP), which replaces $\sigma_{U}(\mathcal{L})$ with an ordering by proximity $\sigma_{P}(\mathcal{L})$, starting from a random node $v_\text{start}\sim\mathcal{U}(\mathcal{L})$:
\begin{equation}
    \sigma_P(\mathcal{L})(v)=\Big|\big\{w\in\mathcal{L}:d(v_\text{start},w)\leq d(v_\text{start},v)\}\Big|.
\end{equation}
 In the case where some nodes are assigned the same index, i.e., when they have exactly the same distance from the starting node, a random order among them is sampled. The aim of the HP ordering is to guide generation from the first node of a layer outwards.

\subsection{Hyperparameters}
\label{sec:appendix_hyperparameters}
We provide a summary of the main hyperparameters of the proposed model in Table~\ref{tab:hparams_shared}. The full list of hyperparameters can be found as Hydra configuration files in the code repository.
% Hyperparameter table for the paper, trimmed to the hyperparameters referenced in the text
% (Secs. 4.1-4.4, 6). Values = what the paper runs used (wandb run configs, all seeds,
% projects scene-graph-gen-midi2_iclr / scene-graph-gen-midi). Config paths are relative to
% the repo's config/ directory. Needs \usepackage{booktabs}.
% MiDi+Or loss weights are the EFFECTIVE ones: the preset says lambda_train=[5,0.4,1,2,0]
% (pos,X,c,E,y order inherited from MiDi) but src/models/midi/metrics/train_metrics.py:56
% applies lambda[2]=1 to the edge loss.
% Notation follows the paper: S = diffusion steps (Eq. 4), lambda_x/e/r (Eq. 6), D = block sizes.
\providecommand{\cfg}[1]{\texttt{\detokenize{#1}}}
 
\begin{table}[t]
\centering
\caption{Hyperparameters of Ours (FLAGG with a MiDi-based Fill module) and of the MiDi+Or
baseline, identical for all datasets. Presets: \cfg{preset/final/midi2/base.yaml} (Ours) and
\cfg{preset/final/midi/base.yaml} (MiDi+Or), relative to \cfg{config/}.}
\label{tab:hparams_shared}
\small
\begin{tabular}{ll}
\toprule
Hyperparameter & Value \\
\midrule
\multicolumn{2}{l}{\textit{Ours --- Insertion and Halting models}} \\
Architecture & GINE, 5 layers, 128 hidden \\
\midrule
\multicolumn{2}{l}{\textit{Ours --- node removal process}} \\
Node ordering $\sigma$ & scene-graph proximity permutation \\
Node schedule & categorical removals, class order $(0,4,5,3,2,1)$ \\
Block sizes $\mathcal{D}$ & $\{1,4\}$ \\
Noisy blocks & block-size noise $p=0.3$ (uniform) \\
Weighted subsampling & $\sqrt{\cdot}$-weighted sampling of $t$, subsampling by weight mass, keep $t{=}1$ \\
\midrule
\multicolumn{2}{l}{\textit{Ours --- Fill module (MiDi-based denoiser)}} \\
Architecture & graph transformer with positions, 9 layers, 8 heads \\
Conditioning GINE & 1 layer, 256 hidden, for each graph transformer layer\\
Diffusion & $S=500$, cosine-$\nu$ schedule, $\nu_X=1$, $\nu_E=1.5$, $\nu_{R}=2.5$ \\
Noise processes & marginal discrete ($\mathbf{X}$, $\mathbf{E}$), VP Gaussian ($\mathbf{R}$) \\
Loss weights (Eq.~\ref{eq:midi_loss}) & $\lambda_x=1$, $\lambda_e=2$, $\lambda_r=4$ \\
Min-SNR & $\gamma=5$ on the coordinate term \\
\midrule
\multicolumn{2}{l}{\textit{Ours --- training and sampling}} \\
Noise augmentation & $p_{\max}=0.5$ (1k-step warm-up); node flip 0.01, edge drop 0.05, \\
 & \quad edge flip 0.025, edge add 0.01, position noise $\sigma=0.1$ \\
Model selection & best validation FGW, checkpoint every 250 steps \\
Sampling batch & 200 graphs \\
\midrule
\multicolumn{2}{l}{\textit{MiDi+Or}} \\
Architecture & graph transformer, 9 layers, 8 heads \\
Diffusion & $S=500$, cosine schedule, marginal transitions, $\nu_X=1$, $\nu_E=1.5$, $\nu_{R}=2.5$ \\
Loss weights (Eq.~\ref{eq:midi_loss}) & $\lambda_x=0.4$, $\lambda_e=1$, $\lambda_r=5$ \\
Model selection & as Ours \\
\midrule
\multicolumn{2}{l}{\textit{Evaluation (both methods; FGW also drives model selection)}} \\
FGW trade-off & $\alpha=0.5$ (structure vs.\ features) \\
\bottomrule
\end{tabular}
\end{table}

\section{Training techniques}
\label{sec:appendix_training_techniques}

\subsection{Mitigating exposure bias}

\paragraph{Noise augmentation.} During training, FLAGG is trained in a teacher-forcing regime, i.e., receiving ground-truth conditioning graphs, while at generation time $G_{t-1}$ contains predictions from earlier steps, which might differ from the training data. We do observe exposure bias in our preliminary tests, leading to worse generation quality in floor and building datasets. Thus, we perturb the training-time conditioning graphs to simulate this effect. The observed plane nodes are protected; previously generated high-level nodes are not. For each non-protected node $v$, we replace its label with another label in $\concepts\setminus\{\textit{plane},\lambda(v)\}$ with probability $p_{\mathrm{node}}$, and add Gaussian position noise with scale $\sigma_{\mathrm{pos}}$:
\begin{equation}
    \widetilde\rho(v)=\rho(v)+\bm\epsilon_v,
    \qquad \bm\epsilon_v\sim\mathcal{N}(\bm0,\sigma_{\mathrm{pos}}^2\bm I_3).
\end{equation}
where $\bm I_3$ is the $3\times3$ identity matrix.
For every node pair with at least one non-protected endpoint, we remove an existing edge with probability $p_{\mathrm{drop}}$ and insert a missing edge with probability $p_{\mathrm{add}}$. We apply this augmentation before encoding $G_{t-1}$ and before MiDi centering; the clean new block remains the prediction target.

\paragraph{Suboptimal categorical removals.} The categorical removal process defined in FLAGG~\citep{cognolato2026flagg} starts from a fixed set of block sizes $\mathcal{D}=\{d_1,\dots,d_l\}$, and computes the optimal combination of sizes to compose the number of nodes of a data graph $G$, representing the blocks in which $G$ can be decomposed. This is solved optimally with a dynamic programming algorithm, but we observe suboptimal choices during inference. Thus, we augment the training data by sampling suboptimal, and thus longer, removal sequences. The suboptimal choices start from the optimal one on a data graph $G$ with $n$ nodes, defined as the multiset of block sizes $\mathcal{B}=\{\{b_1,\dots,b_k\}\}$, such that $\sum_{i=1}^k b_i = n$, and $b_i\in\mathcal{D}$. Then, we choose a size $d$ from $\mathcal{D}$ which has $c_d\in\{0,\dots,|\mathcal{B}|\}$ occurrences in $\mathcal{B}$, such that $d(c_d+1)< n$. Then, we compute the optimal decomposition on $n'=n-d (c_d+1)$ into blocks $\mathcal{B}'$ using the dynamic programming algorithm, and add to $\mathcal{B}'$ the $c_d+1$ occurrences of $d$. This gives a new multiset $\mathcal{B}''$ with $\sum_{i=1}^{k'} b_i' = n$, and $|\mathcal{B}''|>|\mathcal{B}|$. This new multiset is chosen with some probability $p_{\mathrm{subopt}}=0.3$ instead of the optimal one, and the removal sequence is computed from it.

\subsection{Exploration and reweighting}

\paragraph{Weighted removal sampling.} The original \ac{FLAGG} paper used full removal sequences as training data, i.e., given a batch of graphs, sample one removal sequence per graph, collate all graphs into a single batch, and use that for training. This might lead to showing trivial graphs multiple times at extreme timesteps, i.e., at the almost complete graph or the almost empty graph, and under-representing the intermediate states where the number of graphs obtained by removing nodes is high. Thus, we restate the batch compilation by sampling removal timesteps $t$ instead of full sequences. Given a data graph $G$, let $q(G_{t}|G_{t+1})$ be the removal process transition, and $q(G_{t}|G_T)$ be the marginal distribution of $G_t$ given $G_T=G$, which can be easily sampled from given the definition of the categorical removals~\citep[see][]{cognolato2026flagg}. $T$ is computed from $G$ as the minimum number of blocks forming $G$ with sizes taken from the fixed set $\mathcal{D}$. Then, let $c_t$ be the number of graphs $G_t$ that can be obtained by removing $T-t$ blocks from $G$. Then, the weight assigned to timestep $t$ is $w_t = \sqrt{c_t}$, and the probability of sampling $t$ is its normalized weight across $T$ timesteps. Furthermore, given a budget $B$ of graphs obtained with this sampling procedure from a batch of $N$ data graphs, we sample $B_i$ graphs from each data graph proportionally to its total weight $W_i=\sum_{t=1}^{T_i}w_t$, with at least 1 graph assigned to all. This gives a bigger slice of training data in the batch to more complex graphs.

\paragraph{Reweighting MSE with Min-SNR.} During preliminary experiments, we observed a drift in generated positions, moving away from the center, and not aligning with the ground truth. After careful inspection of the trajectory taken by coordinates during denoising, we found that at steps close to $0$ (i.e., almost clean samples) the model was only required to output small adjustments, but instead produced large updates. After an isolated training on these timesteps, the model produced the correct adjustments. The problem lay in the imbalance of gradient contributions. We noticed a much higher gradient from noisy samples (i.e., with $s$ close to $S$) than from clean samples (i.e., with $s$ close to $0$). Thus, we implemented the Min-SNR reweighting~\citep{hang2023snr}, which tackles exactly this problem. In our case, we compute weights as $w_\gamma(s)=\min(\operatorname{SNR}(s), \gamma)$, where $\operatorname{SNR}(s)=\alpha_s^2/\sigma_s^2$ is the signal-to-noise ratio at step $s$, and $\gamma$ is a hyperparameter that we set to $5$ as in the original paper.

\section{Factorized Distance and Discrete Denoising Diffusion Model (FD4) as Fill Module}
\label{sec:appendix_fd4}

\begin{figure*}[h]
  \begin{center}
    \centerline{\includegraphics[width=\textwidth]{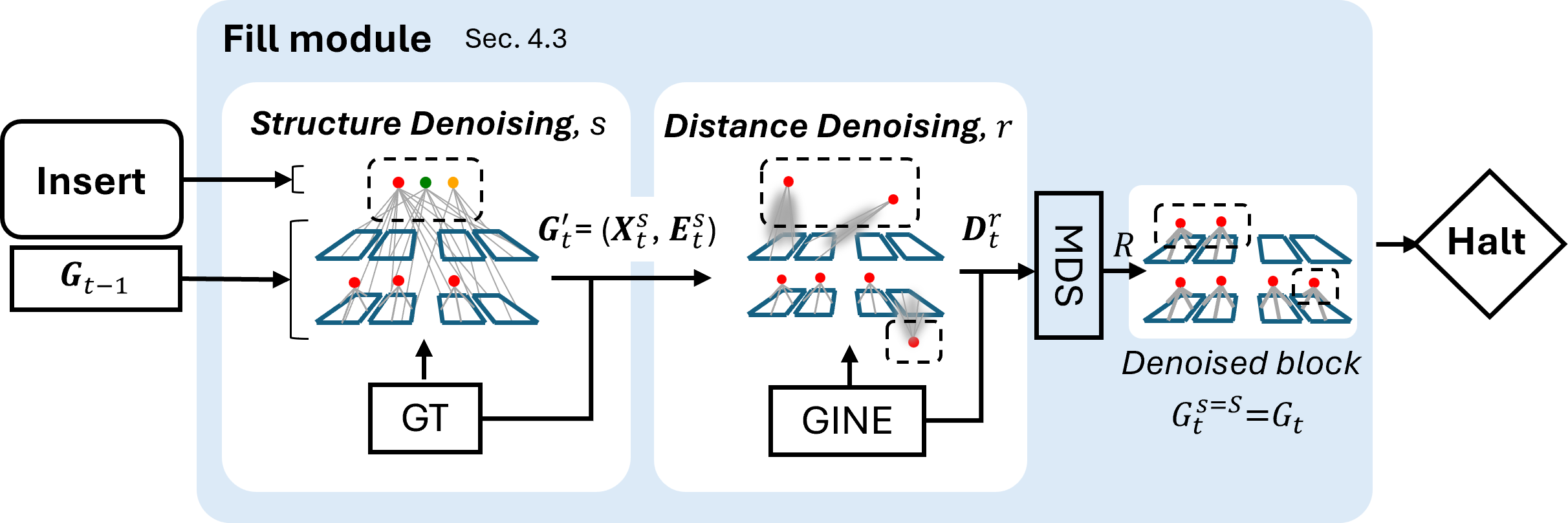}}
    \caption{
      FD4 variant in the Fill module
    }
    \label{fig:architecture_df4}
  \end{center}
\end{figure*}

In this paper we also experiment with another existing one-shot model for 3D graph generation as our Fill module, i.e., \ac{D4}~\citep{cognolato2026d4}.
\ac{D4} natively samples the graph structure and full distance matrix jointly, then recovers 3D coordinates with \ac{MDS}~\citep{torgerson1952multidimensional}. In a 3DSG, the position of a high-level concept depends on the lower-level concepts connected to it. We therefore factorize the adapted model into two learned stages followed by deterministic reconstruction. We call the factorized model FD4. Figure~\ref{fig:architecture_df4} depicts the relevant block using FD4 instead of MiDi in Fig.~\ref{fig:architecture}. First, DiGress~\citep{vignac2023digress} samples $\fX_t$ and $\fE_t$, using the same modeling seen for MiDi, without the 3D coordinate path. Merging the sampled topology with $G_{t-1}$ returns an intermediate graph $G'_t$. Then, a distance model samples the distances on the generated edges. Let $\fD_t\in\R_+^{|\edges_t|-|\edges_{t-1}|}$ contain the distances on all new edges. At distance-denoising step $r$, an edge-regression GINE parameterizes the mean $\mu_\theta(\fD_t^r,G'_t)$, while $\Sigma_r$ is the scheduled covariance. The Gaussian reverse transition is
\begin{equation}\label{eq:distance_transition}
    p_\theta(\fD_t^{r-1}\mid\fD_t^r,G'_t)
    =\mathcal{N}\bigl(\mu_\theta(\fD_t^r,G'_t),\Sigma_r\bigr).
\end{equation}
We then recover $\fR_t$ by minimizing the MDS stress over new-new and new-old edges:
\begin{equation}
    \begin{gathered}
        \fR_t^*=\arg\min_{\fR_t}\sum_{(v,v')\in\edges_t\cap(\verts_t'\times\verts_t')}\left(\|\fR_v-\fR_{v'}\|-\fD_{(v,v')}\right)^2 + \\
        \sum_{(v,v')\in\edges_t\cap(\verts_t'\times\verts_{t-1})}\left(\|\fR_v-\rho(v')\|-\fD_{(v,v')}\right)^2.
    \end{gathered}
\end{equation}
The sampled topology and recovered centroids are merged with $G_{t-1}$ to form $G_t$.

\section{Extended baselines}
\begin{table}[!t]
\centering
% \caption{\textbf{Method comparison.} While the two EC baselines only generate rooms and walls, the random baseline and Ours provide the complete hierarchy (Floor, Building, City). In the architecture, Edge Classification and clustering principles are substituted by autoregressive diffusion.
\caption{\textbf{Method comparison.} \add{The EC baselines are limited to wall/room generation, whereas \textsc{MiDi+Or}, and \textsc{Ours} generate the full hierarchy (floor, building, city).}}
\label{tab:methods}
\setlength{\arraycolsep}{2pt}
\renewcommand{\arraystretch}{1}
\resizebox{\columnwidth}{!}{%
\begin{tabular}{|c|ccc|c|c|cc|ccc|}
\hline
\multicolumn{1}{|c|}{} &
\multicolumn{5}{c|}{\textbf{Generation}} &
\multicolumn{5}{c|}{\textbf{Architecture}} \\ \hline

\textbf{Method} &
\textbf{R} & \textbf{W} & \textbf{F,B,C} & \textbf{Centroids}  & \textbf{Graph size} &
\textbf{EC} & \textbf{Clustering} & \textbf{1-shot} & \textbf{Autoreg.} & \textbf{Diff.} \\ \hline

\textsc{EC-$\square$}           & \cmark$\square$ & \cmark & \xmark & \xmark & \cmark & \cmark & \cmark & \cmark & \xmark & \xmark\\
\textsc{EC-L}                   & \cmark & \cmark & \xmark & \cmark & \cmark & \cmark & \cmark & \cmark & \xmark & \xmark\\
% \textsc{Rand}                   & \cmark & \cmark & \cmark & \cmark & \cmark & \xmark & \xmark & \cmark & \xmark & \xmark\\ 
\textsc{\add{MiDi+Or}}         & \cmark & \cmark & \cmark & \cmark & \xmark & \xmark & \xmark & \cmark & \xmark & \cmark\\ \hline
% \textsc{\add{Ours-FD4}}         & \cmark & \cmark & \cmark & \cmark & \cmark & \xmark & \xmark & \xmark & \cmark & \cmark\\
\textsc{\add{Ours}}         & \cmark & \cmark & \cmark & \cmark & \cmark & \xmark & \xmark & \xmark & \cmark & \cmark\\ \hline
\end{tabular}
}
\end{table}

Please find a full summary of the baselines used in our experiments in Table~\ref{tab:methods}. We provide a brief description of each baseline below.

\subsection{One-shot generation of 3DSGs}
\label{subsec:one_shot_gen}
While our autoregressive model covers the task in all its aspects, it is worthwhile to consider current one-shot approaches in graph generation for comparison. One-shot models receive the final size of the graph as input, usually sampled from the empirical distribution of the dataset, and sample the whole connectivity, labels, and features jointly. A caveat encountered with 3DSG generation is how to implement conditioning: only a new portion of the graph needs generation starting from an input graph, and the size of the former depends on the latter. On the graph side, we resolve the issue by inpainting~\citep{lugmayr2022repaint} the missing nodes, and masking the ground truth plane nodes. For graph sizes, we provide an oracle to the model to make it runnable, i.e., the correct number of nodes is given to the one-shot model. This gives an unfair advantage to the one-shot model compared to our method, but the alternative, i.e., sampling from the empirical distribution, would be an unreliable candidate. We choose MiDi~\citep{vignac2023midi} as the comparison one-shot model.

\subsection{Random baseline}
\label{sec:appendix/base/random}
For our comparison we also considered a random baseline, serving as a weak reference for the metrics. In the following we describe the sampling procedure.
For a dataset, we first compute the training-set empirical distributions of
\begin{enumerate*}[label=(\it\roman*)]
    \item number of non-plane nodes $p_\text{nodes}(n)$;
    \item non-plane labels $p_\text{labels}(l_e)$;
    \item the frequency $p_\text{edges}$ of edges having a non-plane node on at least one end over all possible edges.
\end{enumerate*}
Algorithm~\ref{alg:random_baseline} first samples the extension size from $p_\text{nodes}$. For each new node, \textsc{Rand} independently samples a label from $p_\text{labels}$ and a centroid from $\mathcal{N}(\bm\mu_\planes,\operatorname{diag}(\bm\sigma_\planes^2))$, where $\bm\mu_\planes$ and $\bm\sigma_\planes^2$ are the coordinate-wise mean and variance of the input planes' centroids. It then inserts every candidate edge with at least one new endpoint independently with probability $p_\text{edges}$. Tables~\ref{tab:stacked_three_sets_rand} and~\ref{tab:real_exp_rand} report the results on the synthetic and MSD datasets and on the real dataset, respectively.

\begin{algorithm}[t]
    \caption{Fitting and sampling the \textsc{Rand} baseline}
    \label{alg:random_baseline}
    \begin{algorithmic}[1]
        \REQUIRE Training set $\mathcal{G}_{\mathrm{train}}$; graph of planes $G_\planes=(\verts_\planes,\edges_\planes)$ with centroids $\rho$
        \STATE Estimate $p_\text{nodes}$ from $\{|\verts(G)\setminus\verts_\planes(G)|:G\in\mathcal{G}_{\mathrm{train}}\}$
        \STATE Estimate $p_\text{labels}$ from the labels of all non-plane training nodes
        \STATE Estimate $p_\text{edges}$ as the fraction of candidate training edges with at least one non-plane endpoint that are present
        \STATE Sample $n\sim p_\text{nodes}$ and create a set $\verts_\text{new}$ of $n$ nodes
        \FORALL{$v\in\verts_\text{new}$}
            \STATE Sample $\lambda(v)\sim p_\text{labels}$ and $\rho(v)\sim\mathcal{N}(\bm\mu_\planes,\operatorname{diag}(\bm\sigma_\planes^2))$
        \ENDFOR
        \STATE $\verts\leftarrow\verts_\planes\cup\verts_\text{new}$ and $\edges\leftarrow\edges_\planes$
        \FORALL{$\{u,v\}\subseteq\verts$ such that $\{u,v\}\cap\verts_\text{new}\neq\varnothing$}
            \STATE Insert $\{u,v\}$ into $\edges$ with probability $p_\text{edges}$
        \ENDFOR
        \RETURN $G=(\verts,\edges,\lambda,\rho)$
    \end{algorithmic}
\end{algorithm}
\FloatBarrier

\begin{table*}[h]
\centering
\small
\setlength{\tabcolsep}{3.5pt}
\caption{\textbf{Evaluation results for Rand algorithm in synthetic and MSD datasets.} Values are reported $\times 10^2$ as mean $\pm$ standard deviation over three seeds.}
\label{tab:stacked_three_sets_rand}
\resizebox{\textwidth}{!}{%
\begin{tabular}{l ccccccc ccccccc ccccccc}
\toprule
 
% ===================== SET 1 =====================
\multicolumn{22}{l}{\textbf{Source: Synthetic}} \\
\midrule
\multirow{2}{*}{Method} &
\multicolumn{7}{c}{S-F-$\square$} &
\multicolumn{7}{c}{S-F} &
\multicolumn{7}{c}{S-C} \\
\cmidrule(lr){2-8}\cmidrule(lr){9-15}\cmidrule(lr){16-22}
& FGW $\downarrow$ & GW$\downarrow$ & W$\downarrow$ & Deg.$\downarrow$ & Spec.$\downarrow$ & Clust.$\downarrow$ & GIN$\downarrow$
& FGW $\downarrow$ & GW$\downarrow$ & W$\downarrow$ & Deg.$\downarrow$ & Spec.$\downarrow$ & Clust.$\downarrow$ & GIN$\downarrow$
& FGW $\downarrow$ & GW$\downarrow$ & W$\downarrow$ & Deg.$\downarrow$ & Spec.$\downarrow$ & Clust.$\downarrow$ & GIN$\downarrow$ \\
\midrule
\textsc{Rand} &
\nlvalpm{65.0}{0.0} & \nlvalpm{50.0}{0.0} & \nlvalpm{80.0}{0.0} & \nlvalpm{84.0}{3.0} & \nlvalpm{55.0}{2.0} & \second{\nlvalpm{141.0}{4.0}} & \nlvalpm{69.0}{3.0} &
\nlvalpm{69.0}{2.0} & \nlvalpm{54.0}{2.0} & \nlvalpm{83.0}{3.0} & \nlvalpm{104.0}{1.0} & \nlvalpm{73.0}{1.0} & \second{\nlvalpm{149.0}{3.0}} & \nlvalpm{74.0}{4.0} &
\nlvalpm{65.0}{1.0} & \nlvalpm{55.0}{1.0} & \nlvalpm{75.0}{1.0} & \nlvalpm{143.0}{0.0} & \nlvalpm{123.0}{2.0} & \nlvalpm{160.0}{1.0} & \nlvalpm{95.0}{2.0} \\
\midrule
 
% ===================== SET 2 =====================
\multicolumn{22}{l}{\textbf{Source: MSD}} \\
\midrule
\multirow{2}{*}{Method} &
\multicolumn{7}{c}{M-F} &
\multicolumn{7}{c}{M-B} &
\multicolumn{7}{c}{M-C} \\
\cmidrule(lr){2-8}\cmidrule(lr){9-15}\cmidrule(lr){16-22}
& FGW $\downarrow$ & GW$\downarrow$ & W$\downarrow$ & Deg.$\downarrow$ & Spec.$\downarrow$ & Clust.$\downarrow$ & GIN$\downarrow$
& FGW $\downarrow$ & GW$\downarrow$ & W$\downarrow$ & Deg.$\downarrow$ & Spec.$\downarrow$ & Clust.$\downarrow$ & GIN$\downarrow$
& FGW $\downarrow$ & GW$\downarrow$ & W$\downarrow$ & Deg.$\downarrow$ & Spec.$\downarrow$ & Clust.$\downarrow$ & GIN$\downarrow$ \\
\midrule
\textsc{Rand} &
\nlvalpm{75.0}{0.0} & \nlvalpm{63.0}{0.0} & \nlvalpm{86.0}{0.0} & \nlvalpm{150.0}{1.0} & \nlvalpm{148.0}{1.0} & \second{\nlvalpm{180.0}{1.0}} & \nlvalpm{95.0}{2.0} &
\nlvalpm{74.0}{1.0} & \nlvalpm{64.0}{1.0} & \nlvalpm{85.0}{1.0} & \nlvalpm{157.0}{0.0} & \nlvalpm{183.0}{1.0} & \nlvalpm{157.0}{1.0} & \nlvalpm{103.0}{1.0} &
\nlvalpm{75.0}{0.0} & \nlvalpm{65.0}{0.0} & \nlvalpm{85.0}{0.0} & \nlvalpm{166.0}{0.0} & \nlvalpm{185.0}{0.0} & \nlvalpm{173.0}{1.0} & \nlvalpm{111.0}{1.0} \\
\bottomrule
\end{tabular}
}
\end{table*}

\begin{table}[h]
\centering
\caption{\textbf{Results in real environments for Rand baseline.} Values are reported $\times 10^2$ as mean $\pm$ standard deviation over three seeds.}
\label{tab:real_exp_rand}
\footnotesize
\setlength{\tabcolsep}{2.5pt}
\begin{adjustbox}{max width=\columnwidth}
\begin{tabular}{l ccc ccc ccc}
\toprule
\multicolumn{10}{l}{\textbf{Source: Real}} \\
\midrule
\multirow{2}{*}{Method} &
\multicolumn{3}{c}{R-F} &
\multicolumn{3}{c}{R-B} &
\multicolumn{3}{c}{R-C} \\
\cmidrule(lr){2-4}\cmidrule(lr){5-7}\cmidrule(lr){8-10}
& FGW $\downarrow$ & GW$\downarrow$ & W$\downarrow$
& FGW $\downarrow$ & GW$\downarrow$ & W$\downarrow$
& FGW $\downarrow$ & GW$\downarrow$ & W$\downarrow$ \\
\midrule
\textsc{Rand} &
\nlvalpm{105.0}{5.0} & \nlvalpm{83.0}{3.0} & \nlvalpm{127.0}{6.0}
& \nlvalpm{81.0}{13.0} & \nlvalpm{70.0}{9.0} & \nlvalpm{92.0}{17.0}
& \nlvalpm{92.0}{11.0} & \nlvalpm{78.0}{7.0} & \nlvalpm{106.0}{15.0} \\
\bottomrule
\end{tabular}
\end{adjustbox}
\end{table}

\newpage
\section{Qualitative results}
\label{sec:appendix_qualitative_results}

\begin{figure*}[h!]
  \begin{center}
    \centerline{\includegraphics[width=1.45\textwidth,angle=-90]{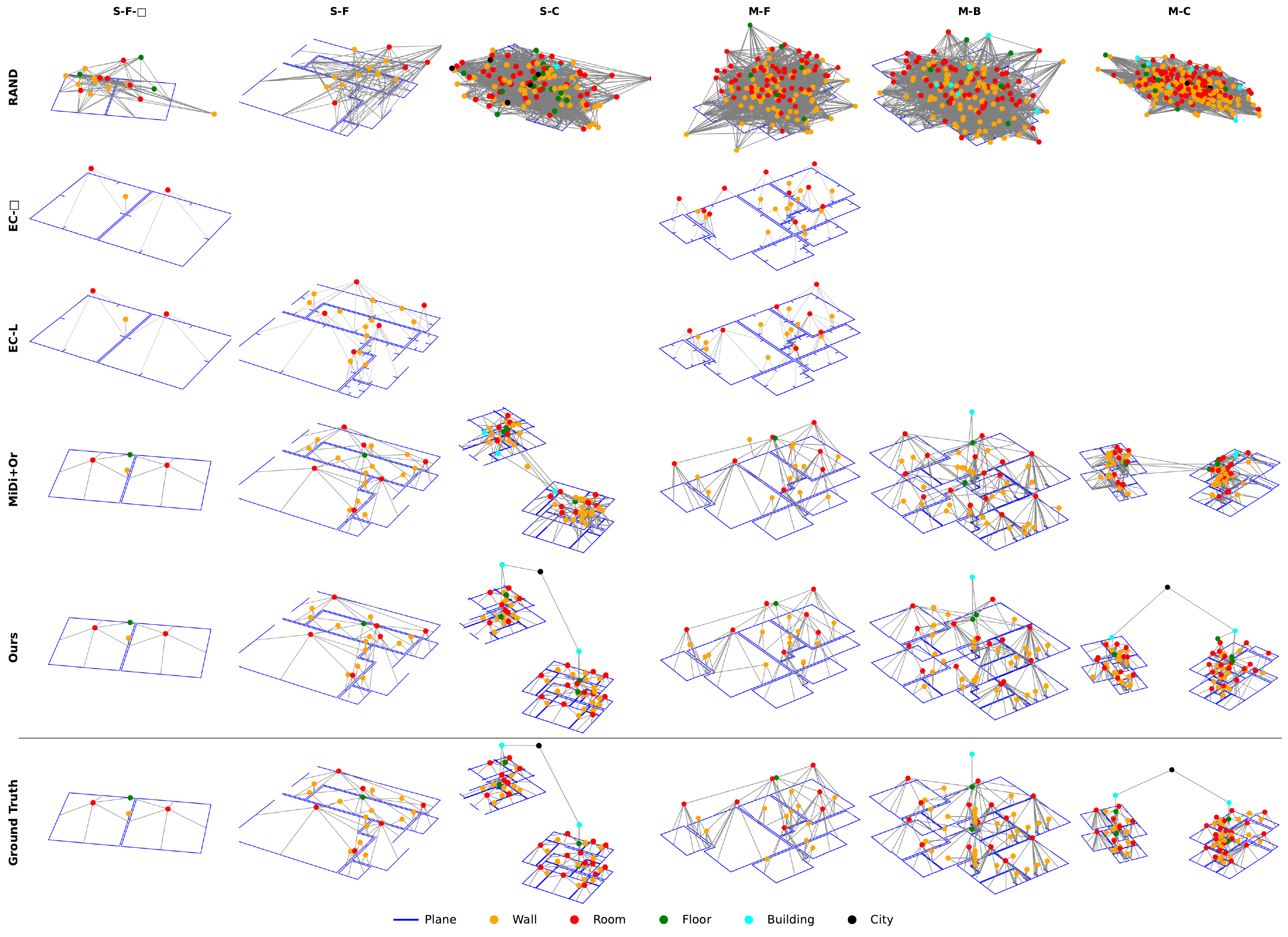}}
    \caption{
      Qualitative results across all training datasets and methods.
    }
    \label{fig:appendix_qualitative_results}
  \end{center}
\end{figure*}

\begin{figure*}[h!]
  \begin{center}
    \centerline{\includegraphics[width=1.0\textwidth,angle=0]{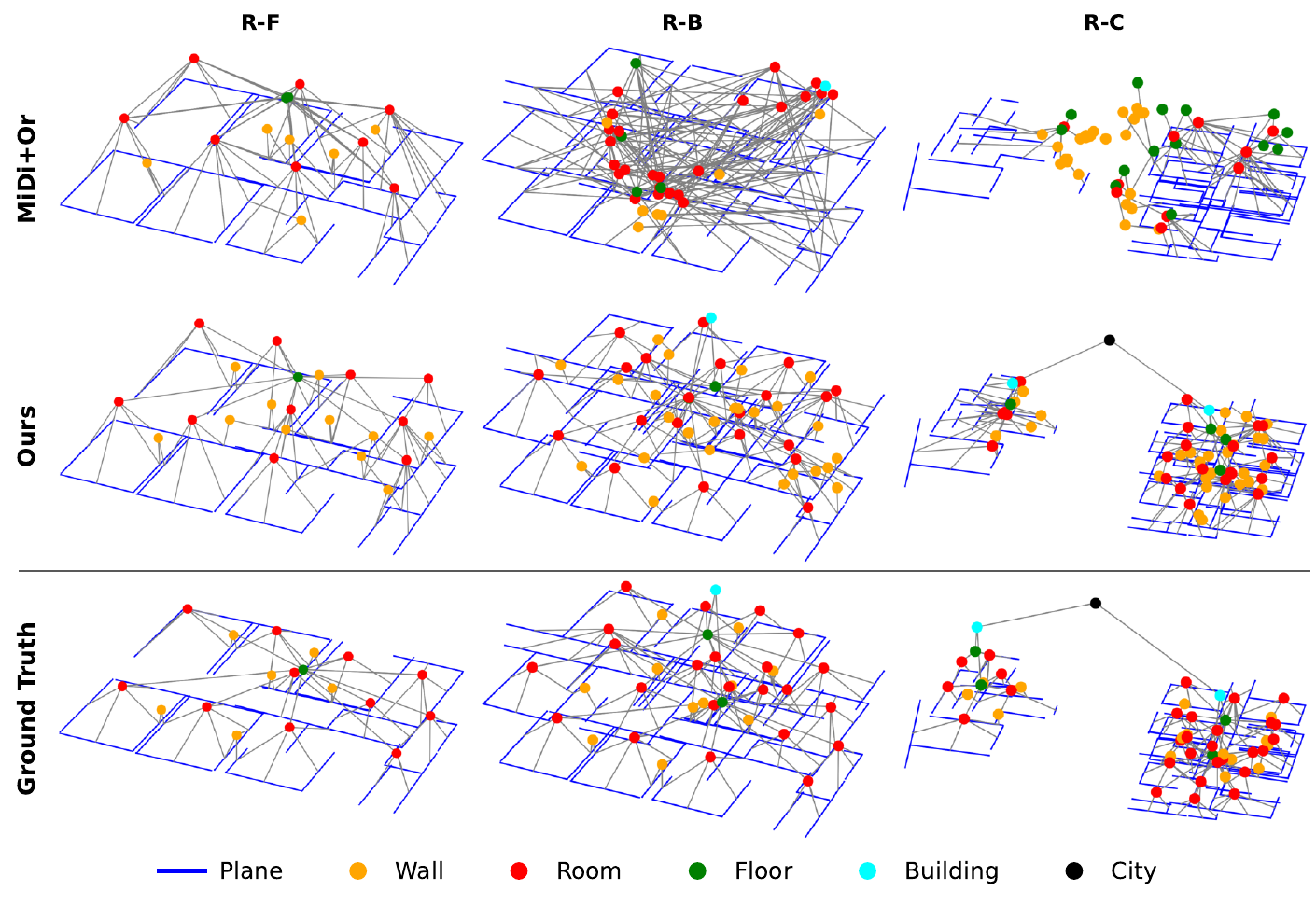}}
    \caption{
      Qualitative results across all real datasets.
    }
    \label{fig:appendix_qualitative_results_real}
  \end{center}
\end{figure*}

\clearpage

\section{Quantitative results}
\label{sec:appendix_quantitative_results}

In Tab.~\ref{tab:sampling_cost}, we report the sampling time and memory for \textsc{MiDi+Or} and \textsc{Ours}. We kept the original implementation of MiDi with minimal adaptations for \textsc{MiDi+Or}, which breaks down the generation of a batch of graphs into sub-batches, trading sampling speed for memory. Additionally, \textsc{MiDi+Or} has access to the true number of nodes, thus saving memory and computation on potentially overflown nodes. This shows in Tab.~\ref{tab:sampling_cost}, as MiDi with oracle has a better memory and time footprint than our model, although \citet{cognolato2026flagg} shows the opposite.

\providecommand{\nlvalpm}[2]{#1\,{\scriptsize$\pm$#2}}
\begin{table}[h]
\centering
\caption{\textbf{Sampling cost on the M- datasets.} Wall-clock time per generated graph and peak GPU
memory while generating from the test split (128 / 200 / 128 graphs), on one NVIDIA V100-32GB,
mean $\pm$ std over 3 seeds. Each method uses its own sampling batch size: \textsc{Ours} generates the whole split in one batch, \textsc{MiDi+Or} in adaptive batches of a few graphs, so peak memory is per configuration, not per graph.}
\label{tab:sampling_cost}
\resizebox{\textwidth}{!}{%
\begin{tabular}{lcccccc}
\toprule
 & \multicolumn{2}{c}{M-F} & \multicolumn{2}{c}{M-B} & \multicolumn{2}{c}{M-C} \\
\cmidrule(lr){2-3} \cmidrule(lr){4-5} \cmidrule(lr){6-7}
Method & Time/graph (s) $\downarrow$ & Mem. (GiB) $\downarrow$ & Time/graph (s) $\downarrow$ & Mem. (GiB) $\downarrow$ & Time/graph (s) $\downarrow$ & Mem. (GiB) $\downarrow$ \\
\midrule
\textsc{MiDi+Or} & \nlvalpm{20.7}{0.8} & \nlvalpm{1.43}{0.00} & \nlvalpm{38.7}{0.5} & \nlvalpm{4.10}{0.00} & \nlvalpm{61.4}{0.5} & \nlvalpm{3.49}{0.00} \\
\textsc{Ours} & \nlvalpm{27.8}{1.1} & \nlvalpm{2.63}{0.02} & \nlvalpm{66.8}{0.9} & \nlvalpm{6.92}{0.04} & \nlvalpm{118.8}{0.4} & \nlvalpm{5.97}{0.03} \\
\bottomrule
\end{tabular}}
\end{table}

\section{Metrics}
\label{sec:app_metrics}
Current graph generation metrics are based on the Maximum Mean Discrepancy (MMD) of a pairwise kernel $k(\cdot,\cdot)$ over graph features, extracted by some function $f$. MMD is defined as:
\begin{equation}
        \operatorname{MMD}^2(p,q) = \mathbb{E}_{G,G'\sim p}[k_f(G,G')]
        +\mathbb{E}_{G,G'\sim q}[k_f(G,G')] - 2\mathbb{E}_{G\sim p,G'\sim q}[k_f(G,G')],
\end{equation}
which can be empirically computed by replacing the expectations with sample averages. The kernel function we use is the Radial Basis Function (RBF) $k(x,x')=\exp(-\gamma\|x-x'\|^2)$. Then, a metric is identified for each feature extractor: the node degree histogram (Deg.), the clustering coefficient histogram (Clust.), the spectrum (Spec.), and finally, embeddings obtained by a randomly initialized Graph Isomorphism Network (GIN)~\citep{xu2018gin}.

\section{Equivariance of the complete model}
\label{sec:appendix_midi_equivariance}
\newtheorem{lemma}{Lemma}

We consider translations and rotations about the vertical axis.  Write such a
rigid transformation as $g=(\bm Q,\bm a)$, where $\bm Q$ is a rotation about
$\bm e_z=(0,0,1)^\top$, and let
$(g\rho)(v)=\bm Q\rho(v)+\bm a$.  Node and edge labels are unchanged, while
plane normals, when present, rotate by $\bm Q$.

Relative to the MiDi architecture~\citep{vignac2023midi}, our Fill model adds
the conditioning graph $G_{t-1}$ and the relative-height feature
\begin{equation}
 h_\rho(v,w)=\left|(\rho(v)-\rho(w))^\top\bm e_z\right|.
 \label{eq:relative_height}
\end{equation}

\begin{lemma}
\label{lem:fill_equivariance}
Assume that $G_{t-1}$ is nonempty and that Eq.~\ref{eq:relative_height} is the
only coordinate-dependent scalar added to the MiDi denoiser.  Then the
conditional Fill kernel is equivariant to translations and rotations about the
vertical axis.  More precisely, let
\begin{equation}
 G_t\sim p^F_\theta(\,\cdot\mid m_t,G_{t-1}),
 \qquad
 \widetilde G_t\sim
 p^F_\theta(\,\cdot\mid m_t,g\mathbin{\cdot}G_{t-1}).
\end{equation}
Then
\begin{equation}
 \widetilde G_t\overset{d}{=}g\mathbin{\cdot}G_t,
 \label{eq:fill_equivariance}
\end{equation}
where $\overset{d}{=}$ denotes equality in distribution.
Its label and edge distributions are invariant, whereas the generated
centroids are equivariant.
\end{lemma}

\begin{proof}
MiDi establishes equivariance of its denoising distribution by centering the
input, using rotation-invariant scalar features, and applying isotropic
Gaussian noise to the coordinates~\citep{vignac2023midi}.  It therefore
suffices to check the two modifications used here: conditioning on
$G_{t-1}$ and adding Eq.~\ref{eq:relative_height}.

First, the anchor defined in Section~\ref{sec:method/filler} is covariant:
\begin{equation}
 \bm c_{t-1}(g\mathbin{\cdot}G_{t-1})
 =\frac{1}{n_{t-1}}\sum_{v\in\verts_{t-1}}
   \bigl(\bm Q\rho(v)+\bm a\bigr)
 =\bm Q\bm c_{t-1}(G_{t-1})+\bm a.
 \label{eq:anchor_covariance}
\end{equation}
Consequently, centering removes the translation and commutes with $\bm Q$:
\begin{equation}
 (g\rho)(v)-\bm c_{t-1}(g\mathbin{\cdot}G_{t-1})
 =\bm Q\bigl(\rho(v)-\bm c_{t-1}(G_{t-1})\bigr).
 \label{eq:center_then_rotate}
\end{equation}

Second, rotations about the vertical axis satisfy
$\bm Q^\top\bm e_z=\bm e_z$.  Therefore the relative height is unchanged:
\begin{equation}
 h_{g\rho}(v,w)
 =\left|\bigl(\bm Q(\rho(v)-\rho(w))\bigr)^\top\bm e_z\right|
 =h_\rho(v,w).
 \label{eq:height_invariance}
\end{equation}
Plane-normal inner products, when used, remain invariant because
$(\bm Q\bm n_v)^\top(\bm Q\bm n_w)=\bm n_v^\top\bm n_w$.  Hence all scalar
inputs remain invariant in the centered frame, so the MiDi argument gives
invariant label and edge distributions and an equivariant coordinate
distribution throughout the reverse process in Eq.~\ref{eq:midi_reverse}.

Finally, let $\bar\rho_t(v)$ be a sampled centroid in the centered frame.  In
the transformed run it is distributed as $\bm Q\bar\rho_t(v)$.  By
Eq.~\ref{eq:anchor_covariance}, restoring the anchor gives
\begin{equation}
 \bm Q\bar\rho_t(v)+\bm c_{t-1}(g\mathbin{\cdot}G_{t-1})
 =\bm Q\bigl(\bar\rho_t(v)+\bm c_{t-1}(G_{t-1})\bigr)+\bm a,
\end{equation}
which is exactly the action of $g$.  This proves
Eq.~\ref{eq:fill_equivariance}.
\end{proof}

\begin{theorem}[Equivariance of the complete generator]
\label{thm:flagg_equivariance}
Assume that $G_\planes$ is nonempty, the Insert and Halt modules use features
invariant to translations and rotations about the vertical axis, and
Eq.~\ref{eq:relative_height} is the only coordinate-dependent scalar added to
the MiDi denoiser.  For a transformation $g$ described above, let
\begin{equation}
 G_\text{final}\sim p_{\phi,\theta,\nu}(\,\cdot\mid G_\planes),
 \qquad
 \widetilde G_\text{final}\sim
 p_{\phi,\theta,\nu}(\,\cdot\mid g\mathbin{\cdot}G_\planes).
\end{equation}
Then
\begin{equation}
 \widetilde G_\text{final}\overset{d}{=}
 g\mathbin{\cdot}G_\text{final},
 \label{eq:flagg_equivariance}
\end{equation}
where $\overset{d}{=}$ denotes equality in distribution.  Thus the complete
FLAGG generator with MiDi as its Fill module is equivariant in distribution
under arbitrary translations and rotations about the vertical axis
($\mathrm{SE}(2)_{xy}\times\R_z$).  Node and edge labels, insertion counts,
and halting decisions are invariant, while centroid coordinates are
equivariant and plane normals rotate with $\bm Q$.  Equivariance is not
guaranteed for rotations that tilt the vertical axis.
\end{theorem}

\begin{proof}[Proof of Theorem~\ref{thm:flagg_equivariance}]
The Insert kernel $p^I_\phi(m_t\mid G_{t-1})$ and Halt kernel
$p^H_\nu(H_t\mid G_t)$ are invariant by assumption.  Conditional on the same
$m_t$, the Fill outputs satisfy the equality in distribution in
Eq.~\ref{eq:fill_equivariance}; merging the sampled block with $G_{t-1}$
commutes with $g$.  Induction over $t$ gives the same law for the halting time
and Eq.~\ref{eq:flagg_equivariance} for the terminal graph.
\end{proof}

The result is not, in general, an $\mathrm{SE}(3)$ statement.  If
$\rho(v)-\rho(w)=\bm e_x$ and $\bm Q$ rotates $\bm e_x$ to $\bm e_z$, the
relative height changes from $0$ to $1$.  Thus the vertical feature restricts
the guaranteed symmetry to translations and rotations about the vertical
axis. Removing this feature would recover the SE(3) symmetry from MiDi.


% \add{\ja{TODO: describe the conditioning-graph noise augmentation and report the selected values of $p_{\mathrm{node}}$, $p_{\mathrm{drop}}$, $p_{\mathrm{add}}$, and $\sigma_{\mathrm{pos}}$.}}

\end{document}